\documentclass[runningheads]{llncs}

\usepackage{eccv}

\usepackage{eccvabbrv}

\usepackage{graphicx}
\usepackage{url}            
\usepackage{booktabs}
\usepackage{mathtools}
\usepackage{wrapfig}
\usepackage{booktabs}
\usepackage{tabularx}
\usepackage{array}
\usepackage{pifont}
\usepackage{multirow}
\usepackage{array}
\usepackage{makecell}
\usepackage{siunitx}
\usepackage{textcomp}
\usepackage[table]{xcolor}

\usepackage[frozencache,cachedir=.]{minted}
\usepackage{xcolor}
\usepackage{listings}
\usepackage{algorithm}
\usepackage{algpseudocode}
\usepackage{caption}
\usepackage{tcolorbox}

\usepackage[accsupp]{axessibility}  

\usepackage{tcolorbox}
\tcbuselibrary{theorems,breakable,skins}

\newcounter{propctr} 

\newtcbtheorem[use counter=propctr]{boxedprop}{Proposition}%
{colback=gray!8,colframe=gray!60,
    fonttitle=\footnotesize\bfseries,
    coltitle=black,
    boxrule=0.6pt,
    arc=2pt,
    left=6pt,right=6pt,top=6pt,bottom=6pt,
    enhanced,
    breakable,
    fontupper=\footnotesize}%
{prop}

\definecolor{lightgraydelta}{HTML}{F0F0F0}
\definecolor{darkgreenacc}{HTML}{1B5E20}
\definecolor{darkredacc}{HTML}{B71C1C}

\definecolor{codebg}{rgb}{0.97,0.97,0.97}

\setminted{
    fontsize=\scriptsize,
    linenos,
    breaklines,
    frame=single,
    tabsize=4,
    bgcolor=codebg,
    style=vs
}

\definecolor{jmincolor}{HTML}{0071bc}
\definecolor{tabcolor}{HTML}{E3F1FA}
\definecolor{tabgray}{HTML}{F2F2F2}
\definecolor{darkgreen}{HTML}{31ed09}
\definecolor{darkergreen}{HTML}{168A00}
\definecolor{lightred}{HTML}{FCE8E8}
\newcommand{\defeq}{\vcentcolon=}

\newcommand{\methodabbr}{FoveateR}
\newcommand{\methodname}{Foveated Reasoner}

\newcommand{\incgreen}{\color{darkergreen}\boldsymbol{\Uparrow}\color{black}}
\newcommand{\decred}{\color{red}\Downarrow\color{black}}

\newcommand{\cmark}{\color{darkgreen}\ding{51}\color{black}}
\newcommand{\xmark}{\color{red}\ding{55}\color{black}}

\usepackage[pagebackref,breaklinks,colorlinks,urlcolor=NavyBlue,citecolor=eccvblue]{hyperref}

\usepackage{orcidlink}

\begin{document}

\title{Foveated Reasoning: Stateful, Action-based Visual Focusing for Vision-Language Models}

\titlerunning{\methodname}

\author{\textbf{Juhong Min\qquad \
Lazar Valkov\qquad \
Vitali Petsiuk\\\vspace{1.2mm}
Hossein Souri\qquad
Deen Dayal Mohan}
\vspace{-0.0mm}}

\authorrunning{J.~Min et al.}

\institute{AI Center -- Mountain View, Samsung Electronics\\
\vspace{1.0mm}
{\small {\tt \url{https://juhongm999.github.io/FoveateR/}}}
\vspace{-4.0mm}}

\maketitle

\begin{abstract}
Vision–language models benefit from high-resolution images, but the increase in visual-token count incurs high compute overhead.
Humans resolve this tension via foveation:
a coarse view guides ``where to look'', while selectively acquired high-acuity evidence refines ``what to think''.
We introduce \methodname, an autoregressive vision-language framework that unifies foveation and reasoning within a single decoding trajectory.
Starting from a low-resolution view, the model triggers foveation only when needed, retrieves high-resolution evidence from selected regions, and injects it back into the same decoding trajectory.
We train the method with a two-stage pipeline: coldstart supervision to bootstrap foveation behavior, followed by reinforcement learning to jointly improve evidence acquisition and task accuracy while discouraging trivial ``see-everything'' solutions.
Experiments show that the method learns effective foveation policies and achieves stronger accuracy under tight visual-token budgets across multiple vision-language benchmarks.
\end{abstract}

\section{Introduction}
\label{sec:introduction}

Large language models (LLMs) have driven substantial progress in natural language processing, and this momentum has naturally extended to vision–language models (VLMs).
Despite good empirical performance, VLMs still face a fundamental bottleneck at the visual front-end:
accuracy typically improves with high-resolution (high-res) input images, but the compute and memory costs grow rapidly as self-attention scales quadratically with the number of visual tokens.
To stay within reasonable resource budgets, recent approaches~\cite{liu2023llava, alayrac2022flamingo, li2023blip2,liu2024llava15} operate on low-resolution (low-res) inputs, sacrificing fine-grained visual details as well as downstream accuracy.
This uncomfortable resolution–efficiency trade-off is particularly problematic in practical deployments where budgets are tight, yet fine-grained visual understanding remains crucial.

\begin{figure}[t]
    \centering
    \scalebox{0.35}{
    \includegraphics{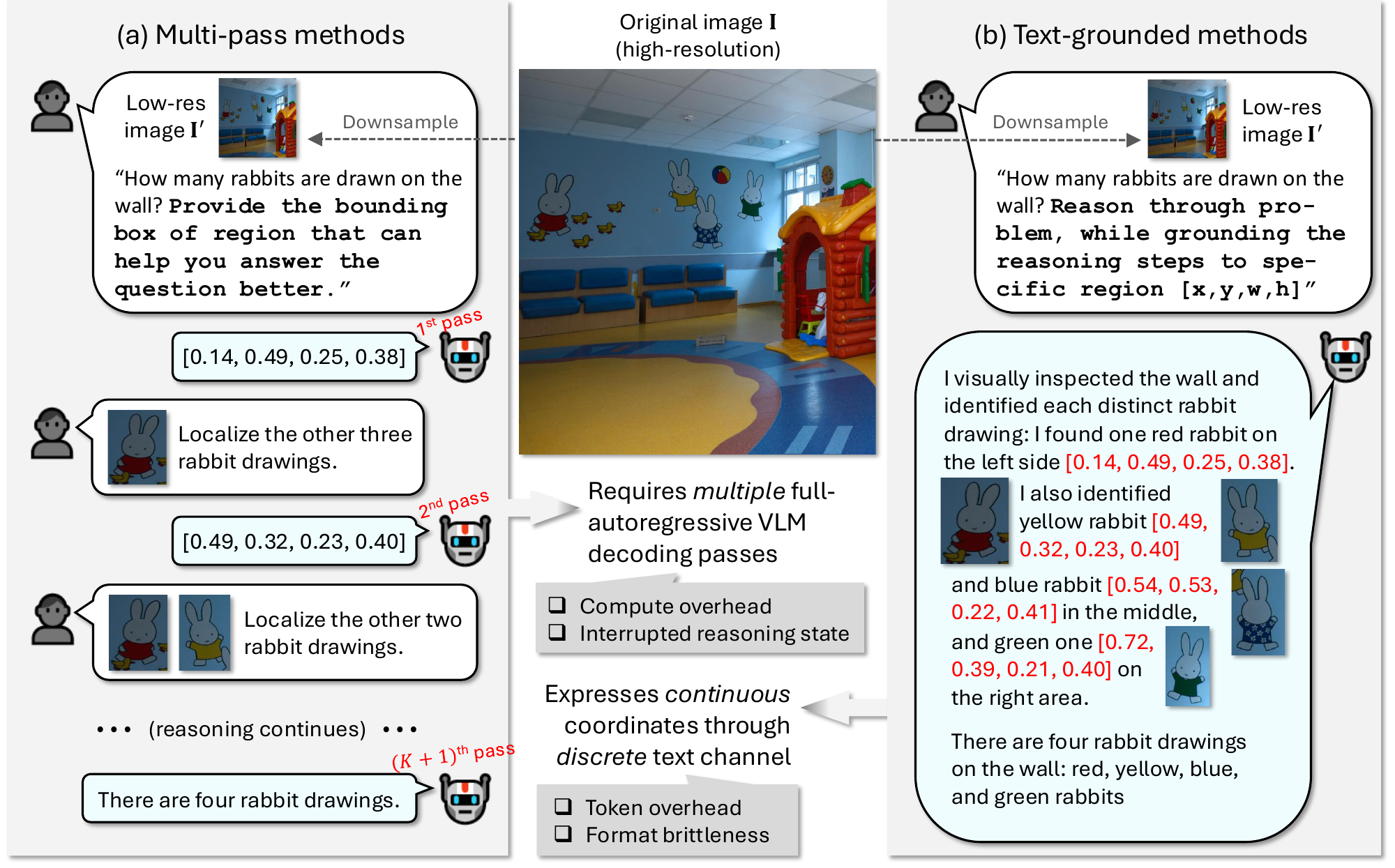}
    }
    \caption{Overview of prior visual focusing methods: (a) multi-pass (left) and (b) text-grounded (right). Given original high-res image (middle), both methods take downsampled image $\mathbf{I}'$ and progressively acquire task-relevant visual evidence from $\mathbf{I}$.}
    \vspace{-7.0mm}
    \label{fig:prior_method_comparison}
\end{figure}

A growing line of work attempts to ease this trade-off through {\em visual focusing}, a strategy that starts with a global low-res view and selectively ``zooms in'' to acquire local high-res evidence from task-relevant regions. 
As shown in Fig.~\ref{fig:prior_method_comparison}, existing visual focusing methods largely fall into two streams:
(a) multi-pass methods~\cite{shao2024visualcot, wu2024vstar, zhao2025uvcot, carvalho2025cropvlm, qi2024cogcom} first run a VLM on a low-res image to decide where to zoom in, then run it again on the high-res crop to answer, often involving multiple zoom-in passes.
(b) text-grounded methods~\cite{zhang2025adaptivecof, fan2025grit, huang2025lavcot, sarch2025vigorl, zheng2025deepeyes, su2025pixelreasoner} represents visual focus signals through free-form texts such as coordinate strings, tool calls, or specialized tokens~\cite{yu2025vpt}, interleaving reasoning with the visual focus in their response.
Although empirically effective, both streams pose practical limitations.
In multi-pass methods, each zoom-in step triggers a new autoregressive run, incurring (i) {\em compute overhead} from multiple decoding passes, and often causing (ii) an {\em interrupted reasoning state}~\cite{shao2024visualcot,zhao2025uvcot} (\ie, re-running the VLM on a new crop breaks hidden-state continuity across passes).
In text-grounded methods, expressing where-to-look coordinates through discrete text channel introduces (iii) {\em token overhead} from repeated coordinate strings and (iv) {\em format brittleness} (\ie, minor formatting deviations can make the output unparsable).
These limitations raise a natural question: What do we miss in current VLM designs to overcome these shortcomings altogether?

One promising answer comes from the human visual system---{\em foveation}~\cite{yarbus2013eye}, in which low-acuity peripheral vision provides low-cost global context to guide where to look, while successive where-to-look decisions (\ie, foveation steps) selectively acquire high-acuity visual evidence.
Importantly, this evidence acquisition process is \textbf{{\em stateful}} and \textbf{{\em action-based}}:
as foveation unfolds, human vision maintains a persistent working memory of previously acquired evidence, thereby preserving an {\em uninterrupted reasoning state} rather than resetting it at each foveation step (as in multi-pass methods).
Moreover, foveation is executed via {\em non-linguistic actions}---eye movements---rather than through explicitly verbalized coordinates (as in text-grounded methods).
This contrast reveals a key gap: while prior methods emulate the {\em outcome} of foveation, they do not instantiate its {\em process}---a stateful, action-based evidence acquisition mechanism.

To bridge this gap, we introduce \methodname~(\methodabbr), an autoregressive VLM pipeline that integrates {\em non-linguistic, action-based} foveation and textual reasoning in a {\em single, stateful} decoding trajectory. 
Starting from a coarse view, \methodabbr~acquires high-res evidence based on the model's evolving hidden state and injects the retrieved evidence into the ongoing generation process.
By preserving an uninterrupted reasoning state and representing foveation as a non-linguistic action, \methodabbr~avoids (i) compute-heavy multiple passes and (ii) hidden state re-initializations, while eliminating both (iii) coordinate-text overhead and (iv) format brittleness, thereby more faithfully capturing key properties of human foveation in VLMs.
Our contributions are threefold:
\begin{enumerate}
    \item We introduce \methodabbr, an autoregressive VLM that integrates action-based foveation with textual reasoning within a single, stateful decoding pass.
    \item We show that our method learns effective foveation policies {\em without human-annotated foveation} trajectories via a two-stage training framework.
    \item Experiments show \methodabbr~outperforms recent visual focusing methods on multiple benchmarks, verifying the efficacy of the proposed approach.
\end{enumerate}

\vspace{-1.0mm}
\section{Preliminaries and Motivations}
\label{sec:motivation}
\vspace{-2.0mm}
In this section, we begin by formalizing standard autoregressive VLMs and their fundamental limitations.
We then discuss two dominant streams of prior visual focusing methods---multi-pass and text-grounded---and their key limitations.
\vspace{-2.0mm}

\subsection{Preliminaries: Autoregressive Vision-Language Modeling} 
We consider a VLM that generates a text response conditioned on a multimodal prompt consisting of an image and an instruction.
Formally, the prompt is denoted by $p \defeq (\mathbf{I}, \mathbf{x})$, where $\mathbf{I} \in \mathbb{R}^{H \times W \times 3}$ is an RGB image and $\mathbf{x} = (x_t)_{t=1}^{|\mathbf{x}|}$ is the input token sequence with $x_t \in \mathcal{V}$ for a finite vocabulary $\mathcal{V}$.
Given $p$, a VLM parameterized by $\theta$ defines an autoregressive conditional distribution over output token sequences.
Specifically, $\pi_\theta(\mathbf{y} \mid p)$ denotes the joint probability of generating the response token sequence $\mathbf{y} = (y_t)_{t=1}^{|\mathbf{y}|}$ conditioned on $p$:
\begin{align}
    \pi_\theta(\mathbf{y} \mid p) = \prod_{t=1}^{|\mathbf{y}|}\pi_\theta(y_t \mid p, y_{<t}),
    \label{eq:standard_vlm}
\end{align}
where $y_{<t}=(y_1, \ldots, y_{t-1})$ denotes the tokens generated before step $t$.

\smallbreak\noindent\textbf{Resolution-efficiency trade-off.}
Typical transformer-based VLMs~\cite{liu2023llava, li2023blip2, alayrac2022flamingo} encode the image $\mathbf{I}$ into a sequence of $N$ visual tokens.
Empirically, increasing the input resolution improves model accuracy~\cite{shao2024visualcot, zhao2025uvcot, jiang2025tokenefficient, shen2025zoomeye, wu2024vstar, carvalho2025cropvlm, qi2024cogcom} but it comes at a steep efficiency cost.
With ViT-style patchification, the token count scales as $N = \Theta(HW)$.
Since self-attention forms an $N\times N$ matrix, the time/memory cost per layer scales quadratically in $N$: $\text{Time/Memory} = \mathcal{O}(N^2) = \Theta((HW)^2)$.
This resolution-efficiency trade-off remains a fundamental limitation of standard VLMs, thus motivating {\em visual focusing} methods to start with low-res inputs and selectively acquire high-res evidence~\cite{shao2024visualcot, wu2024vstar, zhao2025uvcot, shen2025zoomeye, carvalho2025cropvlm, qi2024cogcom, zhang2025adaptivecof, fan2025grit, huang2025lavcot, sarch2025vigorl, zheng2025deepeyes, su2025pixelreasoner, yang2025visionthink, yu2025vpt,lee2026ergo}.

\subsection{Prior visual focusing methods and their limitations}
Prior visual focusing methods can be broadly categorized into two streams: (i) multi-pass~\cite{shao2024visualcot, wu2024vstar, zhao2025uvcot, shen2025zoomeye, carvalho2025cropvlm, qi2024cogcom},
and (ii) text-grounded~\cite{zhang2025adaptivecof, fan2025grit, huang2025lavcot, sarch2025vigorl, zheng2025deepeyes, su2025pixelreasoner, yang2025visionthink, yu2025vpt}.

\smallbreak\noindent\textbf{(i) Multi-pass visual focusing} performs multiple autoregressive (AR) decoding passes to selectively acquire high-res evidence.
Given a downsampled image $\mathbf{I}'=\text{\small{Downsample}}(\mathbf{I})\in\mathbb{R}^{H'\times W'\times 3}$ and a pre-localization prompt $\mathbf{x}^{\mathrm{loc}}$ (\eg, ``\small{\texttt{<Question> Provide the bbox of the region helpful for answering the Question.}}''),
they first run a localization VLM $\pi_{\mathrm{loc}}$ to obtain a bounding box $b$:
\begin{align}
    b=[x,y,w,h] \sim \pi_{\mathrm{loc}}(\cdot \mid p^{\mathrm{loc}}), \qquad
    p^{\mathrm{loc}} \defeq (\mathbf{I}', \mathbf{x}^{\mathrm{loc}}).
\end{align}
They then tokenize the corresponding high-res crop and run an answering VLM $\pi_{\mathrm{ans}}$ conditioned on the high-res evidence $\mathbf{V}$ and the original question $\mathbf{x}$:
\begin{align}
    \mathbf{y} \sim \pi_{\mathrm{ans}}(\cdot \mid p^{\mathrm{ans}}), \qquad
    p^{\mathrm{ans}} \defeq (\mathbf{V}, \mathbf{x}), \qquad \mathbf{V}=\mathrm{Tokenize}(\mathrm{Crop}(\mathbf{I},b)),
\end{align}
Although empirically effective, there remain several limitations.
(1) Most methods~\cite{shao2024visualcot, wu2024vstar, zhao2025uvcot, shen2025zoomeye, carvalho2025cropvlm, qi2024cogcom} perform $K \ge 1$ localization passes which results in $K{+}1$ {\em full autoregressive runs} (including the final answering pass) as formulated below:
\begin{align}
    \underbrace{b_1 \sim \pi_{\mathrm{loc}}(\cdot \mid p^{\mathrm{loc}}_1)}_{\text{pass 1}}
\;\to\; \cdots \;\to\;
\underbrace{b_K \sim \pi_{\mathrm{loc}}(\cdot \mid p^{\mathrm{loc}}_K)}_{\text{pass $K$}}
\;\to\;
\underbrace{\mathbf{y} \sim \pi_\mathrm{ans}(\cdot \mid p^{\mathrm{ans}}(b_{1:K}))}_{\text{final $K$+1 pass}},
\label{eq:reencoding}
\end{align}
thus making multi-step focusing expensive.
(2) As each pass starts a new decoding trajectory, which resets the VLM hidden states, it breaks the continuity of the model's internal reasoning state.
(3) Several systems separate localization from answering ($\pi_{\mathrm{loc}} \neq \pi_{\mathrm{ans}}$)~\cite{wu2024vstar, carvalho2025cropvlm}, improving specialization but increasing total parameters and system complexity.
While reusing the same model for both ($\pi_{\mathrm{loc}} = \pi_{\mathrm{ans}}$)~\cite{shao2024visualcot, shen2025zoomeye, qi2024cogcom, sarch2025vigorl, zhao2025uvcot} is more parameter-efficient, it couples two competing skills (localization {\em vs.} answering) into a single capacity budget, which can hurt either localization or answer quality unless model size is increased.
(4) Some methods~\cite{shao2024visualcot, zhao2025uvcot, carvalho2025cropvlm} use a fixed number of zoom-in steps $K$, rather than dynamically adapting it conditioned on task/question difficulty.

\smallbreak\noindent\textbf{(ii) Text-grounded visual focusing} methods~\cite{fan2025grit,huang2025lavcot,zhang2025adaptivecof,sarch2025vigorl, yu2025vpt} train VLMs to emit spatial references through natural language channel (\eg, coordinate strings such as $[x,y,w,h]$, tool-call text, or augmented localization tokens{\footnote{\cite{yu2025vpt} augments the vocabulary with $\sim$2K localization tokens. While this reduces format brittleness, it still discretizes inherently continuous foveation actions, introducing quantization constraints and extra token-engineering overhead.}).
Importantly, they interleave reasoning with the spatial references (\eg, ``the man is at $(x_1,y_1)$; behind him at $(x_2,y_2)$ there is a ball'').
While conceptually appealing, casting visual focus as free-form text introduces some practical drawbacks:
(1) The approach relies on a \emph{text-to-geometry contract}: coordinates must follow strict conventions (\eg, normalization, coordinate frame, ordering, delimiters, rounding), and deviations can yield non-executable outputs or incorrect region selection.
(2) Learning continuous spatial values through a discrete language channel (\ie, vocabulary set $\mathcal{V}$) quantizes the action space, which can potentially impair fine-grained control over where to zoom in.
(3) Emitting spatial references as output tokens introduces \emph{token overhead}, since coordinates/tool-call arguments consume sequence budget in addition to reasoning tokens.

\smallbreak\noindent\textbf{Motivation for the proposed approach.}
Existing visual focusing methods improve the accuracy--efficiency trade-off, but remain limited by either multi-pass design or text-grounded foveation.
Human foveation is effective not only because it is coarse-to-fine, but also because it performs {\em stateful} evidence acquisition through {\em non-linguistic actions} within a single ongoing process.
Prior methods, however, emulate the selective zoom-in without instantiating this core process, leading to repeated decoding, broken reasoning state, or token overhead.
This motivates a \emph{stateful, single-pass} and \emph{non-linguistic, action-based} visual focusing mechanism, which we present in the next section.

\section{Proposed Method}
\label{sec:method}

We present \methodabbr, a vision-language generation framework that acquires a dynamic amount of high-res visual evidence within a single autoregressive decoding pass.
As shown in Fig.~\ref{fig:overall_pipeline}, we interpret the decoder as an {\em agent} that interacts with a partially observable \emph{environment} (the underlying high-res image).
Based on an internal {\em state} of the agent, it chooses an \emph{action} at each time step:
either emitting the next text token or requesting additional high-res evidence from a foveated region.
We start by formalizing this as a partially observable Markov Decision Process (POMDP).

\subsection{\methodabbr~as a Partially Observable MDP}
\label{sec:pomdp}

We interpret \methodabbr~decoding as a sequential decision process under partial observability:
the agent must answer a question given only a low-res global view, and may acquire (partially observable) local high-res evidence from the original image.

\smallbreak\noindent\textbf{Environment (latent state).}
Each episode starts by providing the agent with $\mathbf{I}'$ and an instruction/question token sequence $\mathbf{x}$, which requests a structured response (\eg, reasoning in \texttt{<think>} and the final answer in \texttt{<answer>}).
The environment also contains the corresponding underlying high-res image $\mathbf{I}$, which remains fixed throughout the episode.
The instruction $\mathbf{x}$ is fully observed, whereas the fine-grained content of $\mathbf{I}$ is only partially observable.
An episode terminates when decoding finishes, \ie, when the model outputs \texttt{<answer>}~$\cdots$~\texttt{</answer>} and then \texttt{<EOS>}.

\smallbreak\noindent\textbf{Observations.}
At each decoding step $t$, the environment returns an observation $o_t$.
The initial observation is the low-res view and the instruction: $o_1=(\mathbf{I}',\mathbf{x})$.
For $t\ge 2$, $o_t$ is either (i) the previously emitted text token $y_{t-1}$ or (ii) a newly revealed high-res evidence.
In case (i), the environment simply echoes the token back to the agent.\footnote{While the ``echoing'' is a non-standard convention as it adds no new external information to the agent, we use it for readability; it makes the autoregressive feedback loop explicit by indicating that each action is conditioned on the previous observations.}

\smallbreak\noindent\textbf{Agent memory.}
The agent maintains a memory buffer $M_t$ that stores the entire history of interactions with the environment (\eg, the initial observation $(\mathbf{I}',\mathbf{x})$, all previously generated text tokens $y_{<t}$, and any high-res evidence revealed from the environment).
In practice, $M_t$ is implemented by the Transformer's KV-cache, which stores the key/value of all previously processed tokens.
The memory is initialized as an empty sequence $M_0=\emptyset$ and updated by concatenating the latest observation:
\begin{align}
    M_{t} = M_{t-1} \oplus o_{t}.
    \label{eq:cache_growing}
\end{align}
As a result, $M_t$ is exactly the input history used to compute the next decoder state.

\begin{figure}[t]
    \centering
    \scalebox{0.4}{
    \includegraphics{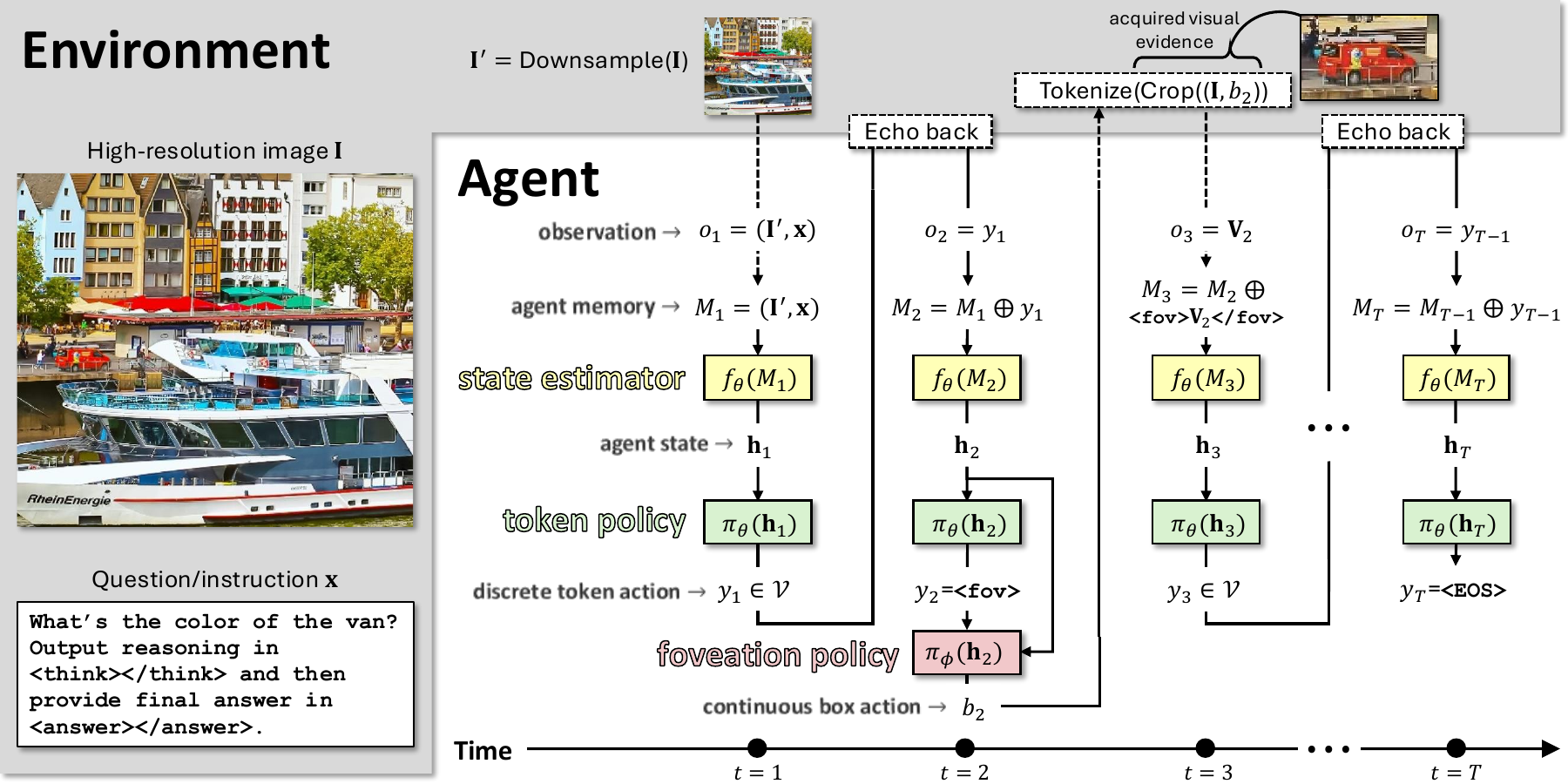}
    }
    \caption{The autoregressive pipeline of the proposed approach where $T = |\mathbf{y}|$.}
    \label{fig:overall_pipeline}
    \vspace{-5.0mm}
\end{figure}

\smallbreak\noindent\textbf{Agent state.}
Since the agent never directly observes the full high-res image $\mathbf{I}$, it must integrate information over time.
We summarize the current memory $M_t$ using the decoder's hidden state $\mathbf{h}_t = f_{\theta}(M_t)\in\mathbb{R}^{d}$ where $f_{\theta}$ is the VLM truncated up to layer $\ell$.
Intuitively, $\mathbf{h}_t$ is the agent's ``belief summary'': a compact summary of what it believes about the unobserved details of $\mathbf{I}$ for the subsequent next action decision.

\smallbreak\noindent\textbf{Actions.}
At each step $t$, the agent selects one of two action modes: emitting the next text token or acquiring additional high-res evidence.
We parameterize these modes with (i) a token policy $\pi_{\theta}$ for discrete token generation and (ii) a foveation policy $\pi_{\phi}$ for continuous region selection, both conditioned on the current agent state $\mathbf{h}_t$.
Concretely, the token policy samples either normal text in $\mathcal{V}$ or special trigger token \texttt{<fov>}:
\begin{align}
y_t \sim \pi_{\theta}(\cdot \mid \mathbf{h}_t), \qquad y_t \in \mathcal{V}\cup\{\texttt{<fov>}\}.
\end{align}
If $y_t\in\mathcal{V}$ (the emitted token is normal text) under our observation convention, the environment echoes it back as $o_{t+1}=y_t$.
If $y_t=\texttt{<fov>}$, the agent instead requests new high-res evidence by sampling a continuous box using the foveation policy,
\begin{align}
b_t \sim \pi_{\phi}(\cdot \mid \mathbf{h}_t), \qquad b_t=[x_t,y_t,w_t,h_t].
\end{align}
Given $b_t$, the environment deterministically returns the corresponding evidence tokens
$\mathbf{V}_t=\mathrm{Tokenize}(\mathrm{Crop}(\mathbf{I},b_t))\in\mathbb{R}^{m_t\times d}$ (where $m_t$ depends on the crop size), which is injected back into the same autoregressive stream as a delimited block \texttt{<fov>}~$\mathbf{V}_t$~\texttt{</fov>} (equivalently, $o_{t+1}=\mathbf{V}_t$), enabling subsequent decoding steps to condition on the newly acquired details without restarting the model.

This design reflects the role separation in human visual cognition: textual reasoning and spatial evidence acquisition are tightly coupled through a shared internal state, yet they operate in different output spaces (discrete linguistic actions {\em vs.}\ continuous geometric actions).
Accordingly, \methodabbr~uses a shared agent state $\mathbf{h}_t$ for communication between reasoning and focusing, while delegating continuous box prediction to a dedicated policy net $\pi_{\phi}$.
Importantly, this head is {\em neither} a separate autoregressive localization model $\pi_{\mathrm{loc}}$ as in multi-pass pipelines, {\em nor} does it express localization through text tokens as in text-grounded methods; 
instead, it is a state-conditioned continuous action predictor executed within the same autoregressive decoding trajectory.

\smallbreak\noindent\textbf{Reward.}
The learning signal combines reward terms for (i) task success from the final answer and (ii) structural correctness of well-formed \texttt{<think>}/\texttt{<answer>} outputs, following~\cite{guo2025deepseekr1}.
During reinforcement learning, these rewards jointly optimize the token policy $\pi_\theta$ and the foveation policy $\pi_{\phi}$.
We refer to Sec.~\ref{sec:training} for details.

\subsection{Interleaved Token-and-Foveation Policy}
Under the partially observable MDP formulation, \methodabbr~follows a joint policy over two types of actions:
it primarily takes {\em discrete token} actions or {\em continuous foveation} actions (only when needed).
Specifically, the backbone token policy $\pi_\theta$ generates the next token $y_t$.
Whenever $y_t=\texttt{<fov>}$, the foveation policy samples a box $b_t\sim \pi_\phi(\cdot\mid\mathbf{h}_t)$ and retrieves evidence $\mathbf{V}_t$ that is injected into the same context for future decoding steps.
We define the set of foveation time steps in the generated sequence as $\mathcal{T} \defeq \{t \mid y_t=\texttt{<fov>}\}$.
\methodabbr~then defines the following joint distribution over the discrete token action sequence $\mathbf{y}$ and the continuous foveation actions $\{b_t\}_{t\in\mathcal{T}}$:
\begin{align}
\pi_{\theta,\phi}\!\left(\mathbf{y}, \{b_t\}_{t\in\mathcal{T}} \mid p\right)
&=
\prod_{t=1}^{|\mathbf{y}|}
\Big[
\pi_\theta\!\left(y_t \mid p, y_{<t}, \{\mathbf{V}_\tau\}_{\tau<t}\right)
\cdot
\pi_\phi\!\left(b_t \mid \mathbf{h}_t\right)^{\mathbb{I}\left[y_t=\texttt{<fov>}\right]}
\Big],
\label{eq:ours_formalization}
\end{align}
where $p\defeq(\mathbf{I}',\mathbf{x})$, $\mathbb{I}[\cdot]$ is indicator function which equals $1$ if its argument is true and $0$ otherwise, and the injected evidence satisfies $\mathbf{V}_t=\mathrm{Tokenize}(\mathrm{Crop}(\mathbf{I},b_t))$ for $t\in\mathcal{T}$.
Eq.~\eqref{eq:ours_formalization} highlights that~\methodabbr~acquires high-res evidence only when triggered within a single autoregressive trajectory, with an adaptive number of foveation steps $|\mathcal{T}|$.

We now relate~\methodabbr~to a standard autoregressive VLM, whose conditional generation distribution is defined in Eq.~\eqref{eq:standard_vlm}.
Intuitively, a standard VLM takes only discrete token actions conditioned on a {\em static visual context}, whereas~\methodabbr~additionally permits ``event-triggered'' continuous foveation actions conditioned on an {\em evolving visual context}.
This relationship can be formalized as Proposition~\ref{prop:ours_reduction}, which shows that standard AR VLMs are a special case of~\methodabbr:
it preserves conventional next-token modeling while enabling {\em optional}, state-dependent evidence acquisition.
This coupling leads to a mixed discrete-continuous learning problem, as both the (discrete) token policy $\pi_\theta$ and the (continuous) foveation policy $\pi_\phi$ must be jointly optimized.
To this end, we employ a two-stage training pipeline that first bootstraps \texttt{<fov>} usage and then uses reinforcement learning to improve task performance as well as foveation behavior.

\begin{boxedprop}{Standard AR VLMs as a Special Case of \methodabbr}{ours_reduction}
    Consider the joint policy of $\pi_{\theta,\phi}$ in Eq.~\eqref{eq:ours_formalization}.
    If the foveation trigger \texttt{<fov>} is never produced, i.e.,
    \begin{align}
    \pi_\theta\!\left(y_t=\texttt{<fov>} \mid p, y_{<t}, \{\mathbf{V}_\tau\}_{\tau<t}\right)=0
    \quad \text{at every decoding step } t,
    \label{eq:no_fov}
    \end{align}
    then~\methodabbr~reduces exactly to the standard autoregressive VLM,
    \begin{align}
    \pi_{\theta,\phi}\!\left(\mathbf{y}, \{b_t\}_{t\in\mathcal{T}} \mid p\right) = \pi_\theta\!\left(\mathbf{y}\mid p\right) = \prod_{t=1}^{|\mathbf{y}|}\pi_\theta(y_t \mid p, y_{<t}).
    \end{align}
    \smallbreak\noindent\emph{Proof.}
    Under Eq.~\eqref{eq:no_fov}, $\mathcal{T}=\emptyset$ for any generated trajectory, so $\mathbb{I}[y_t=\texttt{<fov>}]=0$ for all $t$.
    Hence the foveation factor in Eq.~\eqref{eq:ours_formalization} evaluates to $1$ at every step, and the joint factorization collapses to the standard autoregressive product in Eq.~\eqref{eq:standard_vlm}. \hfill$\square$
\end{boxedprop}

\subsection{Training \methodname}
\label{sec:training}

We initialize \methodabbr~from a pretrained instruction-following VLM (\eg, Qwen2.5-VL~\cite{bai2025qwen25vl}), which provides the state estimator $f_{\theta}$ as well as the token policy $\pi_{\theta}$.
We augment this backbone with a lightweight foveation policy $\pi_{\phi}$, implemented as an MLP with nonlinearities which maps the hidden state to continuous box actions.
Since the pretrained backbone is not trained to use \texttt{<fov>} tags and the foveation policy is newly initialized, we train \methodabbr~in two stages:
(i) coldstart supervised finetuning (SFT) to introduce \texttt{<fov>} usage and box prediction, and (ii) reinforcement learning (RL) to further improve task performance and learn task-effective foveation trajectories.

\smallbreak\noindent\textbf{(1) Coldstart SFT.}
The goal of this stage is to teach a pretrained VLM how to use the foveation mechanism in its decoding stream.
However, standard visual reasoning supervision does not specify when \texttt{<fov>} should be triggered, which region should be queried, or how to integrate retrieved evidence.
Coldstart dataset construction is therefore required to convert such samples into explicit teacher-forcing targets.

\smallbreak\noindent\textbf{Dataset construction.}
Starting from a standard visual reasoning dataset~\cite{shao2024visualcot}, where each sample is a triplet $(\mathbf{I}, \mathbf{x}, \mathbf{a})$ of an image, a question/instruction, and a target answer, we construct a teacher-forcing target sequence $\tilde{\mathbf{y}}=(\tilde{y}_t)_{t=1}^{|\tilde{\mathbf{y}}|}$ that explicitly interleaves reasoning text with foveation segments.
Given $(\mathbf{I}, \mathbf{x})$, a pretrained reasoning VLM~\cite{bai2025qwen25vl} first generates an intermediate reasoning text (\ie, a step-by-step rationale before the final answer), which is then split into $S$ reasoning sentences $\{l_{(s)}\}_{s=1}^{S}$.
For each sentence $l_{(s)}$, we obtain a bounding box $b_{(s)}=\mathrm{Localize}(\mathbf{I}, l_{(s)})$, which localizes the image region described by that sentence.
From each box, we extract its corresponding high-res tokens
$\mathbf{V}^{\star}_{(s)} \defeq \mathrm{Tokenize}(\mathrm{Crop}(\mathbf{I}, b_{(s)}))$.
Finally, we build the following interleaved teacher-forcing target, where foveation and reasoning are interleaved within \texttt{<think>} tags:
\begin{align}
    \tilde{\mathbf{y}} \defeq\;
    &\mathbf{x} \ \texttt{<think>}
    \ \texttt{<fov>}\ \mathbf{V}^{\star}_{(1)}\ \texttt{</fov>}\ l_{(1)}
    \ \cdots\
    \texttt{<fov>}\ \mathbf{V}^{\star}_{(S)}\ \texttt{</fov>}\ l_{(S)} \nonumber\\
    &\texttt{</think>} \ \texttt{<answer>}\ \mathbf{a}\ \texttt{</answer>}.
    \label{eq:coldstart_interleave}
\end{align}
By placing foveation $\mathbf{V}^{\star}_{(s)}$ before each reasoning sentence $l_{(s)}$, the model is trained to first acquire task-relevant visual evidence and then produce reasoning based on that evidence.
This design is consistent with a broader pattern in human cognition, where newly acquired evidence should be incorporated before subsequent reasoning updates.

\smallbreak\noindent\textbf{Training objectives.}
Coldstart SFT jointly supervises (i) next-token generation by $\pi_{\theta}$ and (ii) box prediction by $\pi_{\phi}$.
For next-token prediction, we apply cross-entropy loss only to non-visual tokens in $\tilde{\mathbf{y}}$ (\ie, textual tokens and special tag \texttt{<fov>}), excluding injected visual tokens $\mathbf{V}^{\star}_{(*)}$ because these are not predicted by the model (but deterministically retrieved from the environment given a \texttt{<fov>} trigger and the corresponding box).
For box prediction, we define a set of foveation trigger positions as $\mathcal{T}_{\mathrm{fov}}(\tilde{\mathbf{y}})\defeq \{t \mid \tilde{y}_t=\texttt{<fov>}\}$.
If the \texttt{<fov>} token at position $t\in\mathcal{T}_{\mathrm{fov}}(\tilde{\mathbf{y}})$ corresponds to sentence $l_{(s)}$, we assign the box supervision target $b_t^\star \defeq b_{(s)}$.
The resulting cold-start objective combines token prediction and box regression:
\begin{align}
\mathcal{L}_{\mathrm{cold}}
&=
\underbrace{-\sum_{t\in\mathcal{M}(\tilde{\mathbf{y}})}
\log \pi_{\theta}\!\left(\tilde{y}_t \mid p, \tilde{y}_{<t}, \{\mathbf{V}^{\star}_\tau\}_{\tau<t}\right)
}_{\mathcal{L}_{\mathrm{LM}}}
\;+\;
\underbrace{\lambda_{\mathrm{box}}
\sum_{t\in\mathcal{T}_{\mathrm{fov}}(\tilde{\mathbf{y}})}
\left\| g_\phi(\mathbf{h}_t)-b^{\star}_t \right\|_{1}
}_{\mathcal{L}_{\mathrm{box}}},
\label{eq:coldstart}
\end{align}
where $\mathcal{M}(\tilde{\mathbf{y}})$ denotes the set of non-visual-token positions, and $g_\phi(\mathbf{h}_t)$ is box predicted from the same foveation policy as the conditional box policy $\pi_\phi(\cdot\mid \mathbf{h}_t)$ (\eg, by taking its mean $g_\phi(\mathbf{h}_t)\;\defeq\;\mathbb{E}_{\,b\sim \pi_\phi(\cdot \mid \mathbf{h}_t)}[\,b\,]$).
In short, this objective bootstraps both \texttt{<fov>} usage and box prediction by foveation policy, serving as initialization prior to RL.

\smallbreak\noindent\textbf{(2) RL finetuning with GRPO.}
The coldstart stage provides useful initialization for foveation behavior, but its supervision is not necessarily optimal for end-task performance.
Both the reasoning $l_{(*)}$ and the foveation labels $b^\star_{*}$ are ``pseudo'' labels from a pretrained VLM, which can bias \methodabbr~toward a teacher-induced foveation/reasoning behavior.
In practice, however, the ideal amount of reasoning and foveation is problem-dependent:
some examples can be solved directly from the coarse view without any foveation (\eg, identifying the color of a salient object), whereas others may demand multiple foveations with no intermediate reasoning (\eg, zooming into a few small text regions to read the answer).
To train this problem-adaptive behavior, we further optimize the model using RL.

\smallbreak\noindent\textbf{RL for token policy.}
We optimize the token policy $\pi_\theta$ with a group-relative policy optimization (GRPO) objective, following the work of~\cite{shao2024deepseekmath, guo2025deepseekr1}.
Specifically, for each prompt $p$, we sample a group of $G$ trajectories
$\{\mathbf{y}^{(i)}\}_{i=1}^{G}\sim \pi_{\theta,\phi (\mathrm{old})}(\cdot\mid p)$ and assign a reward $r^{(i)} = r_{\mathrm{acc}}^{(i)} + r_{\mathrm{fmt}}^{(i)}$, where $r_{\mathrm{acc}}^{(i)}=1$ if the extracted final answer is correct, and $r_{\mathrm{fmt}}^{(i)}=1$ if the output follows the required structural format (\eg, well-formed \texttt{<think>} and \texttt{<answer>} tags); otherwise each term is $0$.
Instead of using an external value function, GRPO computes a relative advantage within each sampled group as follows: $A^{(i)} = r^{(i)} - 1/G\sum_{j=1}^{G} r^{(j)}$, centering the reward at the mean, so above-average trajectories obtain positive advantage, while below-average ones are penalized by negative advantage.
We then optimize $\pi_\theta$ using the GRPO objective (omitting PPO-style clipping for brevity):
\begin{align}
\mathcal{L}_{\mathrm{GRPO\text{-}tok}}
\!=\!
-\mathbb{E}\Bigg[
\frac{1}{G}\sum_{i=1}^{G} \frac{1}{|\mathcal M^{(i)}|} \sum_{t \in \mathcal M^{(i)}}
\frac{\pi_{\theta, \phi}(y_{t}^{(i)}\mid p)}{\pi_{\theta, \phi (\mathrm{old})}(y_{t}^{(i)}\mid p)}
A^{(i)}
\!-\!
\beta\,\mathbb{D}_{\mathrm{KL}}
\Bigg],
\label{eq:grpo_token}
\end{align}
where $\mathcal M^{(i)} = \mathcal M(\mathbf y^{(i)})$ is the set of non-visual-token positions in trajectory $i$, $\mathbb{D}_{\mathrm{KL}}$ is a regularization term~\cite{guo2025deepseekr1} that prevents overly large updates of $\pi_{\theta,\phi}$ by keeping it close to the reference model (\eg, coldstart model) and $\beta>0$ controls the regularization.

\smallbreak\noindent\textbf{RL for foveation policy.}
We optimize the foveation policy $\pi_\phi$ with a GRPO objective analogous to the token-policy update (Eq.~\eqref{eq:grpo_token}), but using only \emph{accuracy advantage} $A_{\mathrm{acc}}^{(i)}=r_{\mathrm{acc}}^{(i)}-1/G\sum_{j=1}^{G} r_{\mathrm{acc}}^{(j)}$ because format reward $r_{\mathrm{fmt}}$ is irrelevant to the foveation prediction.
At each foveation step $t\in\mathcal{T}(\mathbf{y}^{(i)})$, the foveation policy samples a box $b_t^{(i)} \sim \pi_\phi(\cdot \mid \mathbf{h}_t^{(i)})$, and we denote the corresponding action likelihood (probability density) by $\pi_\phi(b_t^{(i)} \mid \mathbf{h}_t^{(i)})$.
We then optimize $\pi_\phi$ using a GRPO objective over the sampled foveation actions as follows:
\begin{align}
\mathcal{L}_{\mathrm{GRPO\text{-}fov}}=-\mathbb{E}\Bigg[
\frac{1}{G}\sum_{i=1}^{G}
A_{\mathrm{acc}}^{(i)}
\sum_{t\in\mathcal{T}(\mathbf{y}^{(i)})}
\frac{\pi_{\phi}(b_t^{(i)} \mid \mathbf{h}_t^{(i)})}
{\pi_{\phi(\mathrm{old})}(b_t^{(i)} \mid \mathbf{h}_t^{(i)})}
\Bigg].
\label{eq:grpo_fov}
\end{align}
As a result, the foveation policy is trained to prefer box selections that improve end-task answer accuracy under the current group-relative comparison.

\smallbreak\noindent\textbf{Foveated region regularization.}
Given the two GRPO objectives only (Eqs.~\eqref{eq:grpo_token} \&~\eqref{eq:grpo_fov}), the model may converge to an undesirable trivial solution:
it can increase reward by simply enlarging the foveated box (\eg, increasing its width and height), thereby acquiring larger high-res regions.
In the extreme, it observes the entire high-res image $\mathbf{I}$, which undermines the intended resolution--efficiency trade-off, constituting a form of reward hacking.
To discourage this behavior, we introduce an additional objective that penalizes large foveated regions only when the model prediction is already correct ($r^{(i)}_{\mathrm{acc}}=1$); notably, our foveation actions are {\em differentiable, continuous box parameters} (rather than linguistic coordinates), which allows us to directly regularize the queried area as follows:
\begin{align}
\mathcal{L}_{\mathrm{reg}}
=
\mathbb{E}\Bigg[
\frac{1}{G}\sum_{i=1}^{G}
r_{\mathrm{acc}}^{(i)}
\sum_{t\in\mathcal{T}(\mathbf{y}^{(i)})}
w_t^{(i)} \cdot h_t^{(i)}
\Bigg],
\label{eq:size_reg}
\end{align}
where the product $w_t^{(i)} \cdot h_t^{(i)}$ is the size of the queried region $b_t^{(i)}=[c_t^{x(i)},c_t^{y(i)},w_t^{(i)},h_t^{(i)}]$ (the sampled foveation action at step $t$ in trajectory $i$).
When a prediction is correct ($r_{\mathrm{acc}}^{(i)}=1$), this term penalizes large foveated regions and encourages the model to solve the task with less high-res evidence.
When incorrect ($r_{\mathrm{acc}}^{(i)}=0$), the regularizer is inactive, allowing the model to use larger regions to acquire additional evidence.
Overall, this regularization encourages \methodabbr~to learn accuracy-preserving yet evidence-efficient foveation, rather than trivially enlarging crops to improve reward.

\smallbreak\noindent\textbf{Final objective.}
The RL-stage objective jointly optimizes the token policy (Eq.~\eqref{eq:grpo_token}), the foveation policy (Eq.~\eqref{eq:grpo_fov}), and the foveated-region regularization (Eq.~\eqref{eq:size_reg}):
\begin{align}
\mathcal{L}_{\mathrm{RL}}=
\mathcal{L}_{\mathrm{GRPO\text{-}tok}}
+
\lambda_{\mathrm{fov}}\,\mathcal{L}_{\mathrm{GRPO\text{-}fov}}
+
\lambda_{\mathrm{reg}}\,\mathcal{L}_{\mathrm{reg}},
\label{eq:rl_final}
\end{align}
where $\lambda_{\mathrm{fov}}$ and $\lambda_{\mathrm{reg}}$ balance the contributions of $\mathcal{L}_{\mathrm{GRPO\text{-}fov}}$ and $\mathcal{L}_{\mathrm{reg}}$, respectively.
We provide additional details of the proposed method in the appendix.

\section{Experiments}
\label{sec:experiment}


\subsection{Experimental setup}

\smallbreak\noindent\textbf{Datasets.}
For training and coldstart dataset generation, we use the training splits of the Visual CoT benchmark~\cite{shao2024visualcot} (438K), RefCOCO/+/g~\cite{kazemzadeh2014referitgame} (321K), and ScienceQA~\cite{lu2022scienceqa} (6K), following the recent baseline~\cite{shao2024visualcot}, which yields a total of 765K $(\mathbf{I}, \mathbf{x}, \mathbf{a})$ triplet samples.
For evaluation, we primarily report results on the validation splits of Visual CoT benchmark, which comprises 12 datasets spanning document/text/chart understanding (DocVQA~\cite{mathew2021docvqa}, TextCaps~\cite{sidorov2020textcaps}, TextVQA~\cite{singh2019textvqa}, DUDE~\cite{vanlandeghem2023dude}, SROIE~\cite{huang2019sroie}, InfographicsVQA~\cite{mathew2021infographicvqa}), general VQA (Flickr30k~\cite{plummer2016flickr30k}, Visual7W~\cite{zhu2016visual7w}), relational reasoning (GQA~\cite{hudson2019gqa}, OpenImages~\cite{kuznetsova2020openimages}, VSR~\cite{liu2023vsr}), and fine-grained understanding (CUB~\cite{wah2011cub}), where we evaluate DUDE, SROIE, and Visual7W in a zero-shot setting following prior baselines~\cite{shao2024visualcot,zhao2025uvcot}.
For consistency with~\cite{shao2024visualcot}, we adopt the same GPT-based evaluation metric~\cite{openai}.
We additionally evaluate on V* Bench~\cite{wu2024vstar}, a multiple-choice benchmark designed to probe fine-grained visual perception, including subsets such as direct attribute, GPT4V-hard, and OCR, using the same MCQ evaluation protocol in~\cite{zhao2025uvcot}.

\smallbreak\noindent\textbf{Implementation details.}
We build \methodabbr~on top of Qwen2.5-VL~\cite{bai2025qwen25vl} (3B and 7B) as the backbone VLM, and compare against both recent visual focusing methods~\cite{shao2024visualcot,zhao2025uvcot,yu2025docthinker,yu2025vpt} and standard VLMs~\cite{liu2023llava,lin2023sphinx}. 
We train \methodabbr~in two stages: coldstart for 3 epochs and RL finetuning for 70k steps, using the same batch size of 16 with learning rates of $3e\text{-}5$ and $1e\text{-}6$, respectively.
For RL, we use task-specific rewards: (i) LLM-based string matching for VQA tasks~\cite{shao2024visualcot}, which returns a score in $[0,1]$ based on semantic agreement between the predicted and ground-truth (GT) answers; 
(ii) IoU-based accuracy for grounding tasks~\cite{kazemzadeh2014referitgame}, assigning 1.0 if the IoU between the predicted and GT boxes is $\ge 0.5$ and 0.0 otherwise; 
and (iii) multiple-choice accuracy for ScienceQA~\cite{lu2022scienceqa}, assigning 1.0 if the predicted option matches the GT and 0.0 otherwise.
Additional implementation details are provided in the appendix.

\setlength{\aboverulesep}{0pt}
\setlength{\belowrulesep}{0pt}

\begin{table*}[t]
\centering
\small
\setlength{\tabcolsep}{3.3pt}
\renewcommand{\arraystretch}{1.12}
\caption{Comparison on the Visual CoT benchmark~\cite{shao2024visualcot}. Method subscripts denote visual focusing type: multi-pass ($*_{\textrm{multi}}$), text-grounded ($*_{\textrm{text}}$).
{\scriptsize uses GT box} indicates use of human-annotated boxes for ``visual focusing supervision'', and {\scriptsize visual token cnt} reports the number of visual tokens the model uses, directly impacting inference cost.}
\label{tab:sota_visualcot_comparison}

\resizebox{\textwidth}{!}{
\begin{tabular}{
l
c c
c c c c c c
c c
c c c
c
}
\toprule
\multirow{2}{*}{Method} &
\multirow{2}{*}{\makecell{\vspace{-1.0mm}\scriptsize{uses}\\\vspace{-1.0mm}\scriptsize{GT}\\\vspace{-1.0mm}\scriptsize{box}}} &
\multirow{2}{*}{\makecell{\vspace{-1.0mm}\scriptsize{visual}\\\vspace{-1.0mm}\scriptsize{token}\\\vspace{-1.0mm}\scriptsize{cnt$(\downarrow)$}}} &
\multicolumn{6}{c}{\textbf{Doc/Text/Chart}} &
\multicolumn{2}{c}{\textbf{General}} &
\multicolumn{3}{c}{\textbf{Relation}} &
\multicolumn{1}{c}{\textbf{Fine}} \rule[-1.3ex]{0pt}{4.0ex}\\

\cmidrule(lr){4-9} \cmidrule(lr){10-11} \cmidrule(lr){12-14} \cmidrule(lr){15-15}
& & &
\small{DocV} &
\small{TxtC} &
\small{TxtV} &
\small{Dude*} &
\small{Sroi*} &
\small{Info} &
\small{F30k} &
\small{V7W*} &
\small{GQA} &
\small{OI} &
\small{VSR} &
\small{CUB} \rule[-1.3ex]{0pt}{4.0ex}\\

\rowcolor{tabgray}
\midrule
\multicolumn{5}{c}{$224^2$ input image resolution:} &
\multicolumn{10}{c}{}
\rule[-1.0ex]{0pt}{3.5ex}\\
\midrule

Sphinx$^{13\mathrm{B}}$~\cite{lin2023sphinx}
& {\xmark} & 1734
& 19.8 & 55.1 & 53.2 & 0.00 & 7.1 & 35.2
& 60.7 & 55.8
& 58.4 & 46.7 & 61.3
& 50.5
 \\

VPT$^{2\mathrm{B}}_{\mathrm{text}}$~\cite{yu2025vpt}
& {\cmark} & $\ge$375
& 12.5 & 53.7 & 46.6 & 10.3 & - & -
& \textbf{68.0} & -
& 60.6 & \textbf{84.2} & 65.7
& \textbf{89.2}
 \\

VisCoT$^{7\mathrm{B}}_{\mathrm{multi}}$~\cite{shao2024visualcot}
& {\cmark} & 1,152
& 35.5 & 61.0 & 71.9 & 27.9 & 34.1 & 35.6
& \underline{67.1} & \underline{58.0}
& \underline{61.6} & 83.3 & 68.2
& 55.6
 \\


\rowcolor{tabcolor}  
\methodabbr$^{3\mathrm{B}}$
& {\xmark} & 242.3$_{\pm 168.7}$
& \underline{61.1} & \underline{70.0} & \underline{76.8} & \underline{48.9} & \underline{65.7} & \underline{42.6}
& 57.4 & 52.9
& 56.7 & 77.7 & \underline{69.4}
& 81.7
\rule[-1.0ex]{0pt}{3.7ex}\\


\rowcolor{tabcolor}  
\methodabbr$^{7\mathrm{B}}$
& {\xmark} & 253.8$_{\pm 162.7}$
& \textbf{73.9} & \textbf{76.4} & \textbf{83.5} & \textbf{55.1} & \textbf{73.8} & \textbf{51.2}
& 59.1 & \textbf{58.6}
& \textbf{64.5} & \underline{83.7} & \textbf{74.4}
& \underline{87.4}
\rule[-1.0ex]{0pt}{3.7ex}\\

\rowcolor{tabgray}
\midrule
\multicolumn{5}{c}{$336^2$ input image resolution:} &
\multicolumn{10}{c}{}
\rule[-1.0ex]{0pt}{3.5ex}\\
\midrule

Llava1.5$^{7\mathrm{B}}$~\cite{liu2023llava}
& {\xmark} & 576
& 24.4 & 59.7 & 58.8 & 29.0 & 13.6 & 40.0
& 58.1 & 57.5
& 53.4 & 41.2 & 57.2
& 53.0
 \\

Llava1.5$^{13\mathrm{B}}$~\cite{liu2023llava}
& {\xmark} & 576
& 26.8 & 61.5 & 61.7 & 28.7 & 16.4 & 42.6
& 62.0 & \underline{58.0}
& 57.1 & 41.3 & 59.0
& 57.3
 \\

UV-CoT$^{7\mathrm{B}}_{\mathrm{multi}}$~\cite{zhao2025uvcot}
& {\xmark} & 1,152
& 28.3 & - & 71.1 & 25.3 & 22.7 & 19.8
& 64.9 & 45.5
& 56.8 & - & 55.3
& -
 \\

VisCoT$^{7\mathrm{B}}_{\mathrm{multi}}$~\cite{shao2024visualcot}
& {\cmark} & 1,152
& 47.6 & 67.5 & 77.5 & 38.6 & 47.0 & 32.4
& \underline{66.8} & 55.8
& \underline{63.1} & \underline{82.2} & 61.4
& 55.9
 \\

DocThink$^{3\mathrm{B}}$~\cite{yu2025docthinker}
& {\cmark} & -
& 46.0 & 66.3 & 74.6 & 21.3 & 48.6 & 35.5
& 66.4 & 57.2
& 48.6 & 48.5 & 62.5
& -
 \\

DocThink$^{7\mathrm{B}}$~\cite{yu2025docthinker}
& {\cmark} & -
& 57.9 & 68.2 & 80.2 & 40.8 & 49.5 & 34.7
& \textbf{67.4} & \underline{58.0}
& 54.6 & 54.2 & 65.6
& -
 \\



\rowcolor{tabcolor}  
\methodabbr$^{3\mathrm{B}}$ (ours)
& {\xmark} & 307.0$_{\pm 156.0}$
& \underline{74.0} & \underline{74.0} & \underline{83.0} & \underline{58.2} & \underline{75.4} & \underline{49.0}
& 60.1 & 56.1
& 60.5 & 77.6 & \underline{68.1}
& \underline{82.7}
\rule[-1.3ex]{0pt}{4.0ex}\\


\rowcolor{tabcolor}  
\methodabbr$^{7\mathrm{B}}$ (ours)
& {\xmark} & 322.5$_{\pm 162.7}$
& \textbf{83.3} & \textbf{78.3} & \textbf{86.2} & \textbf{62.7} & \textbf{83.2} & \textbf{60.0}
& {60.8} & \textbf{61.3}
& \textbf{68.2} & \textbf{85.5} & \textbf{73.9}
& \textbf{88.6}
\rule[-1.3ex]{0pt}{4.0ex}\\

\bottomrule
\end{tabular}
}
\vspace{-6.0mm}

\end{table*}

\setlength{\aboverulesep}{0pt}
\setlength{\belowrulesep}{0pt}

\begin{wraptable}{r}{0.46\textwidth}
\centering
\footnotesize
\setlength{\tabcolsep}{6pt}
\renewcommand{\arraystretch}{1.10}
\vspace{-1.0mm}
\caption{Zero-shot results on V*~\cite{wu2024vstar}. Results of~\cite{yao2024minicpm,2024omnilmm,zhao2025uvcot,shao2024visualcot} are from~\cite{zhao2025uvcot}.}
\label{tab:sota_vstar_comparison}
\resizebox{\linewidth}{!}{%
\begin{tabular}{l c c c}
\toprule
{Method ($336^2$ input)} & {Attr} & {GPT4} & {OCR} \\

\midrule

MiniCPM$^{\mathrm{8B}}$~\cite{yao2024minicpm}
& 32.2 & 11.8 & 10.0 \rule[-1.0ex]{0pt}{3.7ex}\\
OmniLMM$^{\mathrm{12B}}$~\cite{2024omnilmm}
& 32.6 & 11.8 & 16.7 \rule[-1.0ex]{0pt}{3.7ex}\\
VisCoT$^{\mathrm{7B}}_{\mathrm{multi}}$~\cite{shao2024visualcot}
& 33.0 & 11.8 & 59.3 \rule[-1.0ex]{0pt}{3.7ex}\\
UV-CoT$^{\mathrm{7B}}_{\mathrm{multi}}$~\cite{zhao2025uvcot}
& 35.2 & 17.6 & 67.7 \rule[-1.0ex]{0pt}{3.7ex}\\
\rowcolor{tabcolor}  
\methodabbr$^{\mathrm{3B}}$ (ours)
& \textbf{56.5} & \textbf{76.5} & \textbf{80.0}
\rule[-1.0ex]{0pt}{3.7ex}\\


\bottomrule
\end{tabular}

}
\vspace{-1pt}
\end{wraptable}

\subsection{Experimental results and analyses}
\vspace{-3.0mm}
\textbf{Comparison with previous approaches.}
Table~\ref{tab:sota_visualcot_comparison} compares \methodabbr\ against recent visual focusing methods~\cite{shao2024visualcot,zhao2025uvcot,yu2025vpt} and standard VLM baselines~\cite{lin2023sphinx,liu2023llava,yu2025docthinker} under two low-res settings, $H',W' \in \{224,336\}$.
Across both 3B and 7B backbones, \methodabbr\ outperforms prior methods on most benchmarks while consuming fewer visual tokens.
Specifically under the $336^2$ setting, our \methodabbr$^{3B}$ (i) uses only $\sim$307 visual tokens on average ({\em vs.}\ 1,152 in~\cite{shao2024visualcot}) (ii) {\em without} using GT-box supervision for foveation policy training, while outperforming GT-trained prior visual focusing methods~\cite{shao2024visualcot,yu2025vpt} on every document understanding task.
We attribute these gains to our RL objectives: $\mathcal{L}_{\mathrm{GRPO\text{-}fov}}$ learns a \emph{goal-driven} foveation policy that directly optimizes end-task accuracy---rather than mimicking GT-box targets---thereby selecting regions most useful for the final answer, while the foveated-region regularizer $\mathcal{L}_{\mathrm{reg}}$ penalizes large foveations to reduce the number of acquired visual tokens.

Since \methodabbr\ adaptively varies its visual budget across samples, we report both the mean and standard deviation of the visual token count (\eg, $307.0_{\pm 156.0}$ for 3B at $336^2$).
The {\em huge variance} is expected: \methodabbr\ uses little or no extra evidence for easy cases, while allocating more foveations to visually demanding cases as shown in Fig.~\ref{fig:qual_comp}.
This problem-dependent allocation is learned during the RL stage (Eqs.~\eqref{eq:grpo_token}-\eqref{eq:size_reg}): by rewarding end-task success, the policy is encouraged to request additional evidence only when it improves the final answer, and to stop foveating otherwise, achieving a favorable accuracy-efficiency balance in expectation.

From Tab.~\ref{tab:sota_vstar_comparison}, we observe a similar trend on V* Bench~\cite{wu2024vstar}, which features substantially higher-resolution images (average $2246\!\times\!1582$) than the Visual CoT benchmarks.
Even in this visually demanding regime, our 3B model under the $336^2$ setting outperforms prior 7-12B methods, highlighting that the proposed evidence acquisition effectively scales to higher-res inputs by foveating only task-relevant regions.

\setlength{\aboverulesep}{0.5pt}
\setlength{\belowrulesep}{0.5pt}

\begin{table*}[t]
\centering
\small
\setlength{\tabcolsep}{3.0pt}
\renewcommand{\arraystretch}{1.12}
\caption{Resolution-efficiency ablation. From a low-res baseline (b), we quantify how it recovers accuracy via selective foveation. Oracle is Qwen2.5-VL with original res.}
\label{tab:ablation}

\resizebox{\textwidth}{!}{
\begin{tabular}{l c c c c c c c c c c c c c c}
\toprule

\multirow{2}{*}{Model} &
\multirow{2}{*}{\makecell{\vspace{-1.0mm}\scriptsize{input}\\\vspace{-1.0mm}\scriptsize{img res.}}} &
\multirow{2}{*}{$\lambda_{\mathrm{reg}}$} &
\multicolumn{4}{c}{DocV~\cite{mathew2021docvqa} (\textbf{1787$\times$2101})} &
\multicolumn{4}{c}{TxtV~\cite{singh2019textvqa} (\textbf{941$\times$820})} &
\multicolumn{4}{c}{V7W~\cite{zhu2016visual7w} (\textbf{556$\times$455})} \rule[-1.3ex]{0pt}{4.0ex}\\

\cmidrule(lr){4-7} \cmidrule(lr){8-11} \cmidrule(lr){12-15}
& & &
$N_{\mathrm{fov}}$ & $\rho$ & $T_{\mathrm{fov}}$ & acc &
$N_{\mathrm{fov}}$ & $\rho$ & $T_{\mathrm{fov}}$ & acc &
$N_{\mathrm{fov}}$ & $\rho$ & $T_{\mathrm{fov}}$ & acc
\rule[-1.3ex]{0pt}{4.0ex}\\

\midrule

(a) oracle$^{\mathrm{7B}}$~\cite{bai2025qwen25vl}
& org & -
& \multicolumn{1}{|c}{-} & 100\% & - & 90.4
& \multicolumn{1}{|c}{-} & 100\% & - & 87.8
& \multicolumn{1}{|c}{-} & 100\% & - & 48.2
\\

\midrule
\midrule

(b) baseline$^{\mathrm{7B}}$~\cite{bai2025qwen25vl}
& $336^2$ & -
& \multicolumn{1}{|c}{-} & 0\% & - & 55.1
& \multicolumn{1}{|c}{-} & 0\% & - & 77.2
& \multicolumn{1}{|c}{-} & 0\% & - & 46.5
\\

\rowcolor{tabcolor}  
(c) ours$^{\mathrm{7B}}_{\mathrm{cold}}$
& $336^2$ & -
& \multicolumn{1}{|c}{35.9} & 3\% & 2.3 & 64.2
& \multicolumn{1}{|c}{73.9} & 7\% & 1.4 & 82.2
& \multicolumn{1}{|c}{32.2} & 8\% & 1.1 & 58.3
\\

\rowcolor{lightred}  
(d) ours$^{\mathrm{7B}}_{\mathrm{cold+RL}}$
& $336^2$ & 1.0
& \multicolumn{1}{|c}{62.5} & 5\% & 2.2 & 61.5
& \multicolumn{1}{|c}{48.8} & 4\% & 1.5 & 81.2
& \multicolumn{1}{|c}{44.3} & 4\% & 1.3 & 57.2
\\

\rowcolor{lightred}  
(e) ours$^{\mathrm{7B}}_{\mathrm{cold+RL}}$
& $336^2$ & 0.5
& \multicolumn{1}{|c}{269.3} & 22\% & 2.3 & 81.5
& \multicolumn{1}{|c}{218.9} & 18\% & 1.7 & 85.4
& \multicolumn{1}{|c}{61.8} & 15\% & 1.2 & 61.0
\\

\rowcolor{lightred}  
(f) ours$^{\mathrm{7B}}_{\mathrm{cold+RL}}$
& $336^2$ & 0.2
& \multicolumn{1}{|c}{354.8} & 25\% & 2.3 & 83.3
& \multicolumn{1}{|c}{263.5} & 21\% & 1.7 & 86.2
& \multicolumn{1}{|c}{81.0} & 17\% & 1.3 & 61.3
\\

\rowcolor{lightred}  
(g) ours$^{\mathrm{7B}}_{\mathrm{cold+RL}}$
& $336^2$ & 0.0
& \multicolumn{1}{|c}{377.8} & 28\% & 2.4 & 74.8
& \multicolumn{1}{|c}{319.6} & 24\% & 2.0 & 85.0
& \multicolumn{1}{|c}{92.0} & 19\% & 1.6 & 59.3
\\

\bottomrule
\end{tabular}
}
\vspace{-4.0mm}
\end{table*}

\begin{figure}[t]
    \centering
    \scalebox{0.4}{
    \includegraphics{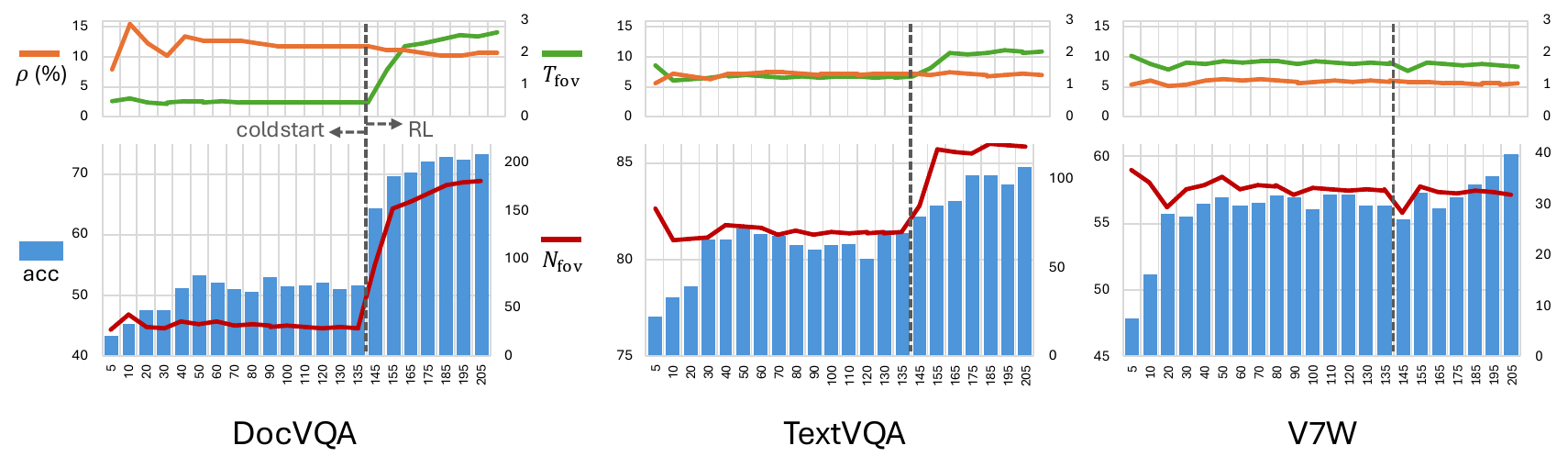}
    }
    \vspace{-3.5mm}
    \caption{Training dynamics of \methodabbr$^{\textrm{3B}}$. The $x$-axis denotes training steps, and the gray vertical line indicates the stage transition from coldstart to RL.}
    \label{fig:training_analysis}
    \vspace{-7.0mm}
\end{figure}

\smallbreak\noindent\textbf{Ablation study.}
Table~\ref{tab:ablation} reports an ablation on our model and baseline~\cite{bai2025qwen25vl}, measuring accuracy with foveation usage statistics---$N_{\mathrm{fov}}$ (additional visual tokens acquired via foveation), $\rho$ (fraction of the high-res image area revealed to the model), and $T_{\mathrm{fov}}$ (number of foveation steps)---on three benchmarks: DocV~\cite{mathew2021docvqa} (high-res documents), TxtV~\cite{singh2019textvqa} (comparatively low-res text-centric images), and V7W~\cite{zhu2016visual7w} (general VQA).
Fig.~\ref{fig:training_analysis} further visualizes the dynamics of these metrics during training.

Comparing (a) with (b), we observe sharp accuracy drops, highlighting aggressive downsampling severely limits fine-grained perception, particularly for text-heavy tasks (\eg, DocV and TxtV).
Comparing the baseline (b) with our coldstart model (c), we observe large gains across all three datasets, even though the model only foveates a small portion of the image ($\rho\!=\!3\%$--$8\%$) with a limited number of extra visual tokens ($N_{\mathrm{fov}}\!=\!32$--$74$, $T_{\mathrm{fov}}\!\approx\!1$--$2$).
This suggests that coldstart is sufficient to bootstrap {\em where-to-look} behavior: reading just a few task-relevant regions often recovers the accuracy or even outperforms the oracle (58.3 {\em vs.} 48.2 on V7W).
The trend is consistent with Fig.~\ref{fig:training_analysis}, where accuracy improves rapidly in initial coldstart (from step 5 to 40) while the foveation usage ($N_{\mathrm{fov}}$, $\rho$, $T_{\mathrm{fov}}$) remains conservative.

Moreover, we study RL stage with different $\lambda_{\mathrm{reg}}$, which penalizes excessive number of visual tokens acquired via foveation (cf. Sec.~\ref{sec:method}):
With a strong regularization ($\lambda_{\mathrm{reg}}\!=\!1.0$, (d)), the model under-utilizes foveation (small $\rho$ and $N_{\mathrm{fov}}$) and fails to improve accuracy.
Relaxing it ($\lambda_{\mathrm{reg}}\!=\!0.5$ and $0.2$, (e)--(f)) allows more foveated evidence ($\rho\!\approx\!15\%$--$25\%$), leading to large accuracy gains on all the three benchmarks.
This is also reflected in Fig.~\ref{fig:training_analysis}: after the coldstart$\rightarrow$RL transition, DocV and TxtV show a sharp increase in $N_{\mathrm{fov}}$ and $T_{\mathrm{fov}}$, with a noticeable jump in accuracy, spending more visual budget when fine-grained reading is necessary.
Removing the regularization entirely ($\lambda_{\mathrm{reg}}\!=\!0.0$, (g)) further increases foveation usage but degrades accuracy, suggesting that {\em unconstrained evidence acquisition} can be inefficient, introducing distractions.
In our experiments, we set $\lambda_{\mathrm{reg}}=0.2$ as a good compromise.

\setlength{\aboverulesep}{0pt}
\setlength{\belowrulesep}{0pt}

\begin{table*}[t]
\centering
\small
\setlength{\tabcolsep}{3.3pt}
\renewcommand{\arraystretch}{1.12}
\caption{Coldstart target ablation under $336^2$ setup. We mark fov-only results with $\incgreen$ ($\decred$) if they are $>2\%p$ higher (lower) than the interleaved counterpart, $\sim$ otherwise.}
\label{tab:fovonly}

\resizebox{\textwidth}{!}{
\begin{tabular}{
l
c
c c c c c c
c c
c c c
c
}
\toprule
\multirow{2}{*}{Model} &
\multirow{2}{*}{\makecell{\vspace{-1.0mm}\scriptsize{cold}\\\vspace{-1.0mm}\scriptsize{start}\\\vspace{-1.0mm}\scriptsize{trg}}} &
\multicolumn{6}{c}{\textbf{Doc/Text/Chart}} &
\multicolumn{2}{c}{\textbf{General}} &
\multicolumn{3}{c}{\textbf{Relation}} &
\multicolumn{1}{c}{\textbf{Fine}} \rule[-1.3ex]{0pt}{4.0ex}\\

\cmidrule(lr){3-8} \cmidrule(lr){9-10} \cmidrule(lr){11-13} \cmidrule(lr){14-14}
& &
\small{DocV} &
\small{TxtC} &
\small{TxtV} &
\small{Dude} &
\small{Sroi} &
\small{Info} &
\small{F30k} &
\small{V7W} &
\small{GQA} &
\small{OI} &
\small{VSR} &
\small{CUB} \rule[-1.3ex]{0pt}{4.0ex}\\

\midrule


(a) ours$^{3\mathrm{B}}_{\mathrm{cold}}$
& $\tilde{\mathbf{y}}$
& 55.4 & 71.6 & 81.4 & 38.4 & 57.8 & 46.0
& 59.3 & 55.5
& 61.2 & 78.9 & 65.1
& 79.5
\rule[-1.3ex]{0pt}{4.0ex}\\


(b) ours$^{3\mathrm{B}}_{\mathrm{cold}}$
& $\tilde{\mathbf{y}}_{\mathrm{fov}}$
& 53.8 ($\sim$) & 72.8 ($\sim$) & 80.8 ($\sim$) & 40.8 ($\incgreen$) & 56.7 ($\sim$) & 43.6 ($\decred$)
& 58.3 ($\sim$) & 57.2 ($\sim$) 
& 64.1 ($\incgreen$) & 81.7 ($\incgreen$) & 67.9 ($\incgreen$)
& 79.5 ($\sim$)
\rule[-1.3ex]{0pt}{4.0ex}\\

\midrule


(c) ours$^{3\mathrm{B}}_{\mathrm{cold+RL}}$
& $\tilde{\mathbf{y}}$
& 74.0 & 74.0 & 83.0 & 58.2 & 75.4 & 49.0
& 60.1 & 56.1
& 60.5 & 77.6 & 68.1
& 82.7
\rule[-1.3ex]{0pt}{4.0ex}\\


(d) ours$^{3\mathrm{B}}_{\mathrm{cold+RL}}$
& $\tilde{\mathbf{y}}_{\mathrm{fov}}$
& 77.3 ($\incgreen$) & 75.9 ($\sim$) & 85.6 ($\incgreen$) & 58.5 ($\sim$) & 74.4 ($\sim$)& 47.9 ($\sim$)
& 59.3 ($\sim$) & 56.9 ($\sim$)
& 62.5 ($\sim$) & 71.4 ($\decred$) & 71.7 ($\incgreen$)
& 84.6 ($\sim$)
\rule[-1.3ex]{0pt}{4.0ex}\\

\bottomrule
\end{tabular}
}

\end{table*}

\begin{figure}[t]
    \centering
    \scalebox{0.35}{
    \includegraphics{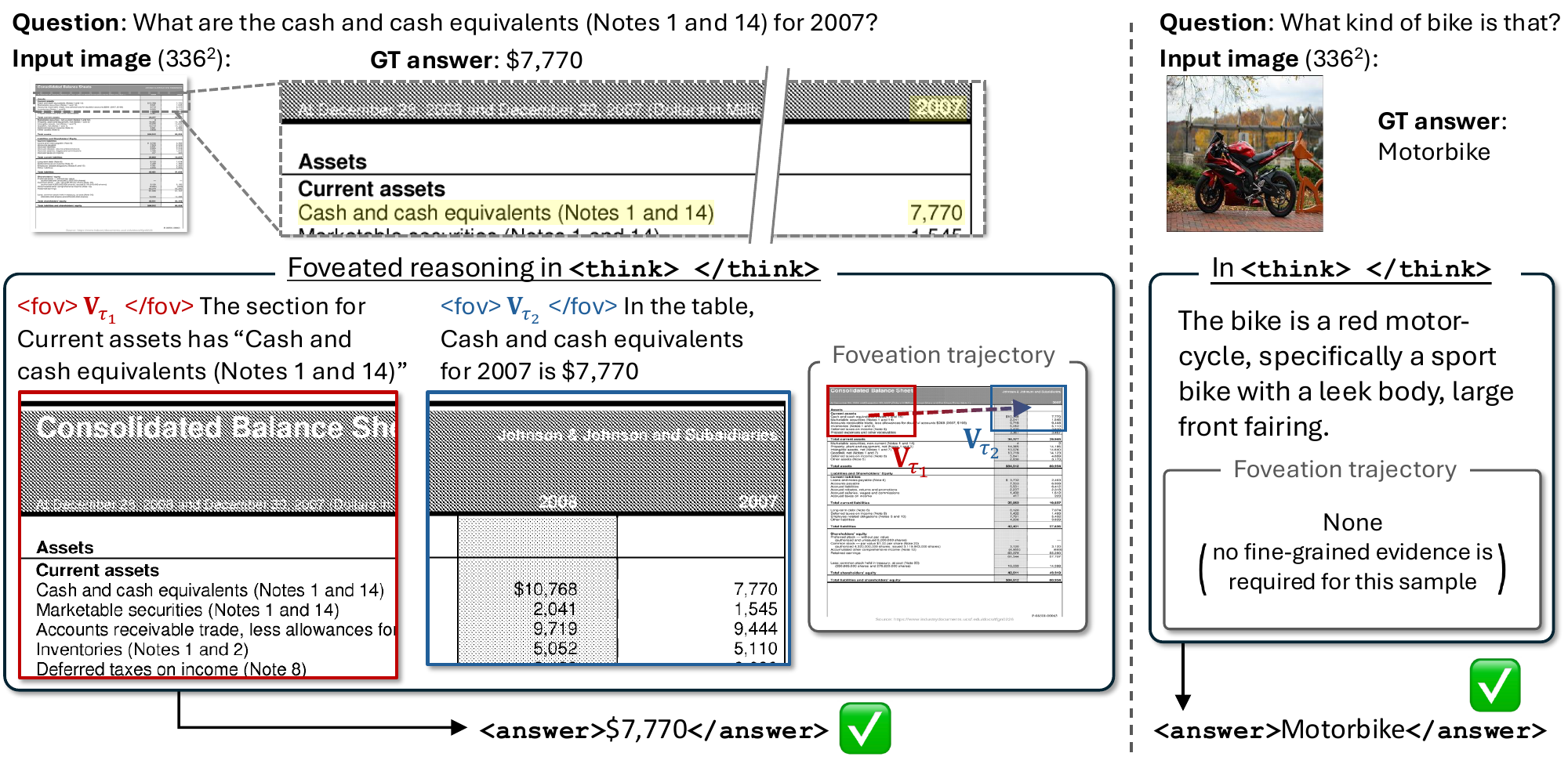}
    }
    \vspace{-3.0mm}
    \caption{Visually demanding cases (left; \eg, documents; more foveations/$N_{\mathrm{vis}}$) {\em vs.}\ less demanding cases (right; \eg, centered dominant objects; fewer or no foveations/$N_{\mathrm{vis}}$).}
    \label{fig:qual_comp}
    \vspace{-6.5mm}
\end{figure}


\smallbreak\noindent\textbf{Coldstart training with `foveation-only' {\em vs.} `interleaved foveation-reasoning'.}
Table~\ref{tab:fovonly} compares two coldstart target constructions:
The first is an \emph{interleaved} target $\tilde{\mathbf{y}}$ that alternates foveated evidences $\mathbf{V}^{\star}_{(*)}$ and rationales $l_{(*)}$ (Eq.~\eqref{eq:coldstart_interleave}).
The second is a {\em foveation-only} (fov-only) target $\tilde{\mathbf{y}}_{\mathrm{fov}}$ that removes the intermediate rationales $l_{(*)}$ and supervises only the visual evidence $\mathbf{V}^{\star}_{(*)}$ acquisition behavior as follows:
\begin{align}
\tilde{\mathbf{y}}_{\mathrm{fov}} \defeq 
\mathbf{x}\ \texttt{<think>}
\texttt{<fov>}\mathbf{V}^{\star}_{(1)}\texttt{</fov>}
\cdots
\texttt{<fov>}\mathbf{V}^{\star}_{(S)}\texttt{</fov>}
\texttt{</think>}\cdots.
\label{eq:coldstart_fovonly}
\end{align}
Interestingly, the two coldstart variants, (a) and (b), achieve comparable performance across most benchmarks, suggesting that coldstart can learn an effective {\em where-to-look} policy even without intermediate narration $l_{(*)}$.
Notably, the fov-only target (b) is even slightly stronger on relation datasets (GQA/OI/VSR).
We attribute this to the nature of their questions (\eg, ``who is dressed in blue?'' or ``what is he wearing on his hand?''), which typically require only {\em one decisive region lookup}, making $\tilde{\mathbf{y}}_{\mathrm{fov}}$ a more direct signal for evidence acquisition than relying on extraneous narration.

We further apply RL to both checkpoints, yielding (c) and (d).
Even after RL, the fov-only checkpoint (d) performs comparably---and often better---than the interleaved checkpoint (c) on most datasets, implying accurate prediction does not necessarily require explicit rationales $l_{(*)}$.
We hypothesize that, in such low-res input setup, what matters more is {\em acquiring the right visual evidence} $\mathbf{V}^{\star}_{(*)}$ through foveation:
intermediate rationales can be redundant, often paraphrasing what is already visible without adding much additional structure, and may even inject noise by encouraging unnecessary narration.
By contrast, learning an effective evidence-acquisition behavior appears more critical under low-res inputs, where success largely depends on capturing decisive high-res visual cues---\emph{a picture is worth a thousand words}.

\clearpage

\begin{table}[t]
\centering
\footnotesize
\setlength{\tabcolsep}{4pt}
\caption{\textbf{Foveation localization quality} on the VisCoT benchmark.
$N_\mathrm{fov}$ and Acc.\ are averaged over all 12 VisualCoT datasets~\cite{shao2024visualcot}; coverage and
IoU@0.5 are shown for four representative ones: DocVQA, V7W, GQA, and CUB.}
\vspace{1.0mm}
\label{tab:localization}
\resizebox{\linewidth}{!}{%
\begin{tabular}{l c c c c cccc cccc c}
\toprule
\multirow{2}{*}{Model} & \multirow{2}{*}{\makecell{GT\\box}} & \multirow{2}{*}{$T_\mathrm{fov}$} & \multirow{2}{*}{$N_\mathrm{fov}$} & \multirow{2}{*}{reas.} &
\multicolumn{4}{c}{GT-region coverage (\%) $\uparrow$} &
\multicolumn{4}{c}{IoU@0.5} & \multirow{2}{*}{Acc.} \\
\cmidrule(lr){6-9}\cmidrule(lr){10-13}
& & & & & DocV & V7W & GQA & CUB & DocV & V7W & GQA & CUB & \\
\midrule
VisCoT$^{7B}_\mathrm{multi}$ & \cmark & fixed(1) & 1152 & \xmark &
1.9 & 8.1 & 11.3 & 13.3 & 19.6 & 27.8 & 46.5 & 45.4 & 57.9 \\
FoveateR$^{3B}_\mathrm{cold+RL}$ & \xmark & dynamic & 258.1 & \xmark &
\textbf{11.0} & 59.3 & 65.7 & \textbf{60.0} & 15.2 & 18.3 & 22.1 & 47.0 & \textbf{68.8} \\
FoveateR$^{3B}_\mathrm{cold+RL}$ & \xmark & dynamic & 242.3 & \cmark &
9.6 & \textbf{63.3} & \textbf{68.5} & 58.3 & 15.1 & 18.5 & 22.6 & 44.7 & 68.2 \\
\bottomrule
\end{tabular}}
\end{table}

\begin{table}[t]
\centering
\scriptsize
\setlength{\tabcolsep}{3pt}
\caption{\textbf{End-to-end efficiency on V7W} (single H100). Visual-token and
latency figures are measured on V7W only; throughput is total processed tokens per second.}
\label{tab:efficiency}
\resizebox{\linewidth}{!}{%
\begin{tabular}{l c c c c c c c c c}
\toprule
\multirow{2}{*}{Model} & \multirow{2}{*}{\makecell{Input\\res.}} & \multirow{2}{*}{Fov.?} & \multirow{2}{*}{Reas.?} & \multirow{2}{*}{$T_\mathrm{fov}$} &
\multirow{2}{*}{\makecell{\# foveated\\vis. tokens}} & \multirow{2}{*}{\makecell{\# text\\tokens}} &
\multirow{2}{*}{\makecell{Throughput\\(tok/s)$\uparrow$}} & \multirow{2}{*}{\makecell{VRAM\\(G)$\downarrow$}} &
\multirow{2}{*}{\makecell{Latency\\(s)$\downarrow$}} \\
\\
\midrule
Qwen2.5-VL$^{3B}$ & orig. & \xmark & \xmark & 0 & 0 & 72.9 & 32.7 & 35 & \underline{2.23} \\
VisCoT$^{7B}$ & $336^2$ & \cmark & \xmark & 1 & 576 & 0 & 58.9 & \underline{25} & 9.78 \\
FoveateR$^{3B}_\mathrm{cold+RL}$ & $336^2$ & \cmark & \xmark &
1.3 & 80.6 & 0 & \textbf{76.0} & \textbf{24} & \textbf{1.06} \\
FoveateR$^{3B}_\mathrm{cold+RL}$ & $336^2$ & \cmark & \cmark &
1.2 & 76.6 & 121.9 & \underline{75.8} & \underline{25} & 2.62 \\
\bottomrule
\end{tabular}}
\vspace{-5.0mm}
\end{table}

\subsection{Foveation Quality and End-to-End Efficiency}
\label{sec:efficiency}

\vspace{-1.0mm}
\noindent\textbf{Foveation localization quality.}
A natural concern is whether the learned policy $\pi_\phi$ attends to task-relevant regions, or merely accumulates evidence until the answer is found.
Since VisCoT predicts a single box whereas FoveateR may emit several, a per-box IoU@0.5 is not directly comparable; 
we therefore report \emph{GT-region coverage}---the fraction of GT-region pixels covered by the union of predicted boxes---which measures how much of the GT evidence is collectively acquired and credits FoveateR's ability to recover from an initially imprecise box at later decoding steps.
As shown in Tab.~\ref{tab:localization}, despite a smaller backbone and \emph{no} human-annotated box supervision, FoveateR covers far more GT evidence than VisCoT$^{7B}$ (\eg, $8.1\!\rightarrow\!63.3$ on V7W) with fewer foveated tokens, while improving average accuracy by over $10\%$p.

\smallbreak
\noindent\textbf{End-to-end efficiency.}
To confirm that the gains are not confined to visual-token count, we measure wall-clock latency (from tokenization to final prediction), peak VRAM, and throughput on a single H100 (Tab.~\ref{tab:efficiency}). Relative to the multi-pass VisCoT$^{7B}$, FoveateR cuts visual tokens by $\sim$$7\times$ and latency by $\sim$$9\times$, with lower peak memory and higher throughput at better accuracy---avoiding the repeated re-encoding that makes multi-pass pipelines costly. Moreover, comparing our two variants isolates the cost of emitting text: adding intermediate reasoning alone raises the text-token budget to $121.9$ and roughly doubles latency ($1.06\!\rightarrow\!2.62$s). This indicates that text-grounded methods, which must additionally verbalize box coordinates at every foveation, would incur an even larger budget---a cost FoveateR avoids by acquiring evidence through non-linguistic, continuous actions. Together, these results show that FoveateR's efficiency stems from \emph{acquiring the right evidence rather than more of it}, validating single-pass foveation as a practical path to high-resolution reasoning under tight budgets.

\section{Related Work}
\label{sec:related_work}
Effective visual evidence acquisition has been studied for decades, from classic neural vision~\cite{mnih2014ram} to modern VLMs~\cite{shao2024visualcot}.
Below, we briefly review some recent relevant work.

\smallbreak\noindent\textbf{Token reduction.}
A natural way to reduce visual cost is to start from a dense set of visual tokens and then \emph{select}, \emph{merge}, or \emph{prune} tokens to form a compact representation~\cite{rao2021dynamicvit,liang2022evit,xu2021evovit,kong2022spvit,yin2022adavit,ryoo2022tokenlearner,bolya2023tokenmerging,liu2023revisitingtoken,goyal2020powerbert}.
One key drawback is that token reduction methods mostly compress an ``already-observed'' representation:
if fine-grained details are missing in the initial (often low-res) encoding, token selection alone cannot recover new information.
Feeding a higher-res image can preserve such details, but it makes the initial encoding itself compute-heavy.
Moreover, aggressive pruning may discard task-critical evidence, making performance sensitive to the selection policy/threshold.

\smallbreak\noindent\textbf{Visual reasoning.}
A large class of VLMs perform visual reasoning, in which the model acquires task-relevant information by composing it through {\em textual reasoning about the visual content}.
In practice, these models~\cite{xu2025llavacot,zhang2024llava-reasoner-dpo,zhang2024multimodalchainofthought,huang2023languageisnot,zhu2023minigpt4,alayrac2022flamingo,li2023blip2,dai2023instructblip,liu2024llava15,wang2023cogvlm,qin2025covt,mathew2021docvqa} typically operate in a {\em fixed-context} manner:
the input image is encoded once into a static set of visual tokens, and all subsequent reasoning is conditioned on this unchanged visual context.
As a result, performance is ultimately bounded by the level of detail preserved in the initial visual input;
increasing its resolution may recover fine-grained details but it quickly brings back the resolution–efficiency trade-off.

\smallbreak\noindent\textbf{Visual focusing.}
To overcome the fixed-context bottleneck, visual focusing methods enable the model to adaptively acquire additional evidence during inference, typically via coarse-to-fine procedures that revisit the high-res image.
While empirically effective, there remain fundamental bottlenecks:
(i) Multi-pass pipelines~\cite{shao2024visualcot, wu2024vstar, zhao2025uvcot, shen2025zoomeye, carvalho2025cropvlm, qi2024cogcom,jiang2025tokenefficient,yang2025visionthink,lee2026ergo} can increase inference latency due to repeated re-encoding/re-prompting, often breaking hidden-state continuity across different passes.
(ii) Text-grounded methods~\cite{zhang2025adaptivecof, fan2025grit, huang2025lavcot, sarch2025vigorl, zheng2025deepeyes, su2025pixelreasoner, yang2025visionthink, yu2025vpt} often introduce token overhead and format-brittleness.
In contrast to both streams, FoveateR differs along multiple axes.
(1) It preserves an uninterrupted trajectory via a shared KV-cache instead of resetting hidden states across passes;
(2) acquires all evidence in a \emph{single} run, conditioning each foveation on prior evidence \emph{and} queried positions;
(3) emits continuous, non-linguistic box actions rather than brittle coordinate strings;
and (4) jointly optimizes foveation and reasoning end-to-end without GT-box supervision.
These instantiate the \emph{process} of foveation—stateful, action-based evidence acquisition—rather than merely emulating its \emph{outcome}.

\smallbreak\noindent\textbf{Foveation in machine vision.}
Early work by Mnih \etal~\cite{mnih2014ram} explored a human-inspired foveation paradigm, where an agent sequentially selects glimpses and integrates them through an internal state.
Although conceptually appealing, it was difficult to demonstrate as a practical solution at the time, due to limited model capacity, unstable optimization, and the lack of large-scale data~\cite{liu2023llava,alayrac2022flamingo}.
Subsequent machine vision research has adopted idea of foveation more broadly, \eg, foveated/peripheral representations~\cite{rosenholtz2012summary,jonnalagadda2022foveater,lukanov2021biologically}, eccentricity-aware pooling~\cite{ye2024neural,Lin_2025}, and peripheral-aware attention~\cite{min2022pervit}.
In this paper, we also draw inspiration from three key characteristics of foveation:
(i) {\em coarse-to-fine} evidence acquisition mechanism, (ii) an {\em uninterrupted, stateful} inference process, and (iii) {\em continuous action-based} control.
Based on these principles, we instantiate foveation as a practical vision–language system.

\vspace{-2.0mm}
\section{Conclusion}
\label{sec:conclusion}
This paper argues that the resolution–efficiency trade-off requires not only ``zooming in,'' but also stateful, action-driven evidence acquisition.
We presented \methodabbr, a stateful, single-pass autoregressive framework that triggers non-linguistic, continuous foveation actions on demand and integrates the retrieved high-resolution evidence into the on-going decoding stream, preserving hidden-state continuity.
Trained with a two-stage pipeline, it learns task-adaptive foveation strategies and delivers strong performance with limited visual tokens on a broad set of vision–language benchmarks. We hope this work inspires future multimodal systems under strict compute budgets.

\setcounter{table}{0}
\setcounter{figure}{0}
\setcounter{equation}{0}
\renewcommand{\thetable}{A\arabic{table}}
\renewcommand{\thefigure}{A\arabic{figure}}
\renewcommand{\theequation}{A\arabic{equation}}
\renewcommand{\thealgorithm}{A\arabic{algorithm}}

\appendix

\noindent\textbf{{\LARGE Appendix}}
\smallbreak\smallbreak\smallbreak\smallbreak\smallbreak\smallbreak

\noindent We provide additional details on the model architecture (Sec.~\ref{sec:supp_architecture}), training (Sec.~\ref{sec:supp_training}), inference (Sec.~\ref{sec:supp_inference}), and coldstart dataset statistics (Sec.~\ref{sec:supp_coldstart}).
We further include ablation studies on foveation policy (Sec.~\ref{sec:supp_abl_foveation}) and coldstart supervision signals (Sec.~\ref{sec:supp_abl_supervision}), along with additional qualitative results and analyses (Sec.~\ref{sec:supp_qual}).
We conclude by discussing the limitations of our method and potential directions for future work (Sec.~\ref{sec:supp_futurework}).

\section{Implementation Details}
\label{sec:supp_framework}

\subsection{Model Architecture}
\label{sec:supp_architecture}

Our model consists of three trainable components: an agent state estimator $f_\theta$, a token policy $\pi_\theta$, and a foveation policy $\pi_\phi$.
The first two components ($f_\theta$ and $\pi_\theta$) are obtained by decomposing a pretrained vision-language backbone (\eg, Qwen2.5-VL-Instruct~\cite{bai2025qwen25vl}) into two parts, where the first $\ell$ transformer layers form $f_\theta$ and the remaining layers form $\pi_\theta$.
The third component, $\pi_\phi$, is a newly initialized MLP that takes the agent state to predict a continuous box action.

\smallbreak\noindent\textbf{Agent state estimator $f_{\theta}$.}
In our implementation, we use the first 27 layers (out of 36 total) of the Qwen2.5-VL~\cite{bai2025qwen25vl} backbone as the agent state estimator $f_{\theta}$, while the remaining 9 layers are used for the token policy $\pi_\theta$.
Empirically, this intermediate depth ($\ell=27$) provided the most effective hidden-state representation $\mathbf{h}_t$ for both the token policy $\pi_{\theta}$ and the foveation policy $\pi_{\phi}$.
We hypothesize that, if the state is extracted from earlier layers ($\ell < 27$), the model has not yet integrated sufficient visual and textual context for reliable foveation decisions;
if it is extracted from later layers ($\ell > 27$), the representation becomes over-specialized for immediate text token prediction, making it less suitable for foveation prediction.
Based on this observation, we set $\ell=27$ in all experiments.

\smallbreak\noindent\textbf{Token policy $\pi_{\theta}$.}
We use the last 9 transformer layers of the backbone~\cite{bai2025qwen25vl} as the token policy $\pi_\theta$, which takes the shared agent state $\mathbf{h}_t$ and predicts the next token $y_t \in \mathcal{V} \cup \{\texttt{<fov>}\}$.
Without the foveation mechanism (\ie, without $\pi_\phi$ and the trigger \texttt{<fov>}), the combination of $f_{\theta}$ and $\pi_{\theta}$ is a standard AR VLM.
When foveation is enabled, however, $\pi_\theta$ is responsible not only for generating textual tokens, but also for selecting between two action modes: continuing textual reasoning ($y_t \in \mathcal{V}$) or triggering a foveation step ($y_t = \texttt{<fov>}$).
In this sense, $\pi_\theta$ determines not only ``what to say next,'' but also ``whether $\mathbf{h}_t$ already has sufficient visual evidence'' for subsequent reasoning or answer prediction.

\smallbreak\noindent\textbf{Foveation policy $\pi_{\phi}$.}
When the token policy emits \texttt{<fov>}, the foveation policy $\pi_\phi$ takes the hidden state $\mathbf{h}_t$ and predicts a continuous box action
\begin{align}
    b_t = [c^x_t, c^y_t, w_t, h_t],
\end{align}
where $(c^x_t, c^y_t)$ denotes the box center and $(w_t, h_t)$ denote its width and height.
Algorithm.~\ref{code:foveation_policy} provides complete (PyTorch-style) pseudocode of the foveation policy used in our implementation.
As shown in the \texttt{forward()} function of Alg.~\ref{code:foveation_policy}, the policy $\pi_\phi$ is implemented as a lightweight two-layer MLP;
it first projects $\mathbf{h}_t$ to $\mathbb{R}^{d/2}$ and then predicts the four continuous box parameters (\texttt{mean\_box}).
To keep the output box bounded, we map the center coordinates using \texttt{tanh()} such that $c^x_t, c^y_t \in (-1,1)$, and map the width and height using a sigmoid followed by clamping such that $w_t, h_t \in [\alpha, 1)$.
In all experiments, we set $\alpha = 0.1$.

The foveation policy is used differently in the coldstart and RL stages.
(i) In the coldstart stage, the policy predicts the box deterministically (\texttt{mean\_box}): $g_\phi(\mathbf{h}_t)\;\defeq\;\mathbb{E}_{\,b\sim \pi_\phi(\cdot \mid \mathbf{h}_t)}[\,b\,]$ (cf. Sec.~\ref{sec:training} and lines 12--23 \& 36--39 in Alg.~\ref{code:foveation_policy}) which is used to compute the box regression loss (Eq.~\eqref{eq:coldstart}).
The same deterministic prediction $g_\phi(\mathbf{h}_t)$ is also used during inference.
(ii) In the RL stage, by contrast, the policy must discover which foveation actions $b_t$ lead to better end-task outcomes.
To enable such exploration, it samples box actions $b_t$ (\texttt{box}) from a Gaussian distribution centered at the current prediction with standard deviation $\sigma=0.1$ (lines 12--35 in Alg.~\ref{code:foveation_policy}).
This exploration encourages the policy to try nearby alternatives, \eg, slightly shifted locations or different box sizes, to discover better foveation behaviors than those learned in the coldstart stage.

\smallbreak\noindent\textbf{Foveation-action log-probability.}
Recall that RL training with GRPO~\cite{shao2024deepseekmath} samples $G$ trajectories, each of which may contain both text-token actions and foveation actions.
To compute the GRPO objectives (Eqs.~\eqref{eq:grpo_token} and~\eqref{eq:grpo_fov}), we require the log-probability of every sampled text-token action, $\pi_{\theta,\phi}(y_{t}^{(i)} \mid p)$\footnote{Note that we write $\pi_{\theta,\phi}$ here because the token trajectory may depend on previously acquired foveated evidence, although the token distribution itself is produced by $\pi_\theta$.}, as well as every sampled foveation action, $\pi_{\phi}(b_{t}^{(i)} \mid \mathbf{h}_t^{(i)})$.
The former, $\pi_{\theta,\phi}(y_{t}^{(i)} \mid p)$, is obtained from the next-token distribution (\ie, logits over the vocabulary) produced by the token policy $\pi_\theta$.
The latter, $\pi_{\phi}(b_{t}^{(i)} \mid \mathbf{h}_t^{(i)})$, is obtained from the foveation policy $\pi_\phi$ conditioned on the corresponding hidden state $\mathbf{h}_t^{(i)}$ as implemented in \texttt{get\_log\_prob()} function of Alg.~\ref{code:foveation_policy}.
Concretely, given $\mathbf{h}_t^{(i)}$ (\texttt{h\_t}), the policy first reconstructs the deterministic box $g_\phi(\mathbf{h}_t^{(i)})$ (\texttt{mean\_box}) and computes how likely the current policy would be to produce $b_t^{(i)}$ (\texttt{box}) from state $\mathbf{h}_t^{(i)}$.
In line 56 of Alg.~\ref{code:foveation_policy}, \texttt{log\_prob} represents $\pi_{\phi}(b_{t}^{(i)} \mid \mathbf{h}_t^{(i)})$: the log-probability of the sampled box action $b_t^{(i)}$ under the current policy.
This log-probability is used to compute $\mathcal{L}_{\mathrm{GRPO}\text{-}\mathrm{fov}}$ in Eq.~\eqref{eq:grpo_fov}, which encourages the policy to assign higher probability to box actions that yield higher accuracy rewards $r_{\mathrm{acc}}$.

\begin{figure}[t]
\captionsetup{type=algorithm}
\centering
\begin{minipage}{0.96\linewidth}
\begin{minted}{python}
class FoveationPolicy:
    def __init__(self, hidden_dim, std=0.1, min_size=0.1, max_size=1.0):
        self.std = std              # std of Gaussian policy used for RL exploration
        self.min_size = min_size    # minimum valid box size
        self.max_size = max_size    # maximum valid box size

        reduced_dim = hidden_dim // 2
        self.proj = nn.Linear(hidden_dim, reduced_dim)  # hidden-state projection
        self.head = nn.Linear(reduced_dim, 4)           # predict 4 box parameters

    def forward(self, h_t, sample=False):
        # h_t: agent state extracted from the truncated backbone f_theta
        feat = F.relu(self.proj(h_t))
        params = self.head(feat)

        # Predict bounded continuous box parameters:
        # center = (c^x, c^y) in [-1, 1], size = (w, h) in [min_size, max_size]
        center = torch.tanh(params[:, :2])            # box center (c^x, c^y)
        size = torch.sigmoid(params[:, 2:4])          # box size (w, h)
        size = torch.clamp(size, self.min_size, self.max_size)

        # Deterministic box prediction g_phi(h_t), used in coldstart and inference
        mean_box = torch.cat([center, size], dim=1)

        if sample:
            # RL stage: sample a box action from a Gaussian centered at mean_box
            # This enables local exploration around the current prediction
            dist = torch.distributions.Normal(mean_box, self.std)
            box = dist.rsample()                      # reparameterized sample for RL
            log_prob = dist.log_prob(box).sum(dim=1)  # log pi_phi(b_t | h_t)

            # Clamp the sampled box to the valid action range before execution
            center = torch.clamp(box[:, :2], -1, 1)
            size = torch.clamp(box[:, 2:4], self.min_size, self.max_size)
            box = torch.cat([center, size], dim=1)   # final valid sampled box action
        else:
            # Coldstart / inference: use the deterministic prediction directly
            box = mean_box
            log_prob = torch.zeros_like(center[:, 0])  # unused outside RL

        return log_prob, box

    def get_log_prob(self, h_t, box):
        # Recompute the current policy distribution from h_t
        # and evaluate how likely the given box action is.
        # This is used in the GRPO objective for sampled foveation actions.
        feat = F.relu(self.proj(h_t))
        params = self.head(feat)

        center = torch.tanh(params[:, :2])
        size = torch.sigmoid(params[:, 2:4])
        size = torch.clamp(size, self.min_size, self.max_size)

        mean_box = torch.cat([center, size], dim=1)   # g_phi(h_t)
        dist = torch.distributions.Normal(mean_box, self.std)
        log_prob = dist.log_prob(box).sum(dim=1)      # log pi_phi(b_t | h_t)
        return log_prob, mean_box
\end{minted}
\end{minipage}
\caption{PyTorch-style pseudocode for the foveation policy \(\pi_\phi\).}
\label{code:foveation_policy}
\end{figure}

\clearpage
\subsection{Training details}
\label{sec:supp_training}

\smallbreak\noindent\textbf{Training datasets.}
We train \methodabbr~on the training splits of three datasets: the Visual CoT benchmark~\cite{shao2024visualcot}, RefCOCO/+/g~\cite{kazemzadeh2014referitgame}, and ScienceQA~\cite{lu2022scienceqa}.
These datasets provide complementary supervision:
Visual CoT serves as the primary training source, as it is the standard benchmark used by recent baseline methods~\cite{shao2024visualcot,zhao2025uvcot,yu2025vpt,yu2025docthinker}, providing the most direct training distribution for comparison with prior work.
RefCOCO/+/g provides grounding-oriented supervision, which improves the model's spatial localization ability and yields an approximately 1-2\%p gain on the Visual CoT benchmarks in our experiments.
ScienceQA is primarily included to expose the model to multiple-choice supervision, which is beneficial for evaluation on another multiple-choice benchmark V* Bench.
Overall, the three datasets complement each other by covering general visual reasoning (Visual CoT), spatial grounding (RefCOCO/+/g), and structured logical multiple-choice reasoning (ScienceQA), resulting in a total of 765K training samples.

\smallbreak\noindent\textbf{Rewards during RL stage.}
During RL training, the accuracy reward $r_{\mathrm{acc}}$ is computed differently depending on the dataset type of each sample.
(1) For VQA-style datasets (\eg, Visual CoT benchmark~\cite{shao2024visualcot}), we use an LLM-based semantic answer matching score using Qwen3-4B~\cite{yang2025qwen3};
during RL stage, we use open source model~\cite{yang2025qwen3}, rather than an API-based evaluator (\eg, GPT), because RL training requires reward computation for a large number of samples, making repeated API calls prohibitively expensive in practice.
For evaluation on the Visual CoT benchmark, however, we use GPT-based answer matching to ensure a fair comparison with prior methods~\cite{shao2024visualcot,zhao2025uvcot,yu2025vpt,yu2025docthinker}.
Given the question, the ground-truth answer, and the model prediction, the evaluator is prompted to return a soft score in $[0,1]$ based on semantic agreement between the prediction and GT.
The prompt template shown below is used for this evaluation, which is adopted from~\cite{shao2024visualcot} to ensure consistency with the evaluation protocol used in prior work~\cite{shao2024visualcot,zhao2025uvcot}.

\begin{tcolorbox}[
    colback=gray!10,
    colframe=black!70,
    boxrule=0.8pt,
    arc=2mm,
    left=2mm,
    right=2mm,
    top=1.5mm,
    bottom=1.5mm
]
    \small
    \textbf{LLM evaluator prompt for VQA-style reward scoring.}
    \medskip
    
    You are responsible for proofreading the answers, you need to give a score to the model's answer by referring to the standard answer, based on the given question. The full score is 1 point and the minimum score is 0 points. Please output the score in the form ``score: <score>''. The evaluation criteria require that the closer the model's answer is to the standard answer, the higher the score.
    \medskip
    
    Question: \{ \}\\
    Standard answer: \{ \}\\
    Model's answer: \{ \}
\end{tcolorbox}
\noindent
(2) For grounding datasets (\eg, RefCOCO/+/g~\cite{kazemzadeh2014referitgame}), we use IoU-based accuracy as the reward.
Given a predicted box $\hat{b}$ and a GT box $b^\star$,
we assign $r_{\mathrm{acc}} = 1.0$ if $\mathrm{IoU}(\hat{b}, b^\star) \geq 0.5$, and $r_{\mathrm{acc}} = 0.0$ otherwise, which directly reflects whether the model has faithfully localized the referred object or region.
(3) For multiple-choice datasets (\eg, ScienceQA~\cite{lu2022scienceqa}), we use multiple-choice option matching:
if the predicted option index matches the ground-truth option index, we assign $r_{\mathrm{acc}}=1.0$; otherwise, $r_{\mathrm{acc}}=0.0$.

\clearpage

\newcommand{\hyperparamtablescale}{1.08}
\setlength{\aboverulesep}{1.0pt}
\setlength{\belowrulesep}{1.0pt}

\begin{table}[t]
\centering
\scriptsize
\setlength{\tabcolsep}{4pt}
\renewcommand{\arraystretch}{1.05}
\caption{Summary of key hyperparameters and training settings.}
\label{tab:hyperparams}
\scalebox{\hyperparamtablescale}{
\begin{tabular}{l l l}
\toprule
\textbf{Group} & \textbf{Hyperparameter} & \textbf{Value} \\
\midrule

\multirow{6}{*}{Architecture}
& Backbone & Qwen2.5-VL-Instruct (3B, 7B) \\
& Layer split ($f_{\theta}$ / $\pi_{\theta}$) & $27 / 9$ \\
& Foveation policy $\pi_{\phi}$ & 2-layer MLP with nonlinearities\\
& Minimum box size $\alpha$ & 0.1 \\
& Box sampling std.\ $\sigma$ & 0.1 \\
& Input resolution $H',W'$ & $224^2$, $336^2$ \\

\midrule
\multirow{10}{*}{Training}
& Datasets & Visual CoT, RefCOCO/+/g, ScienceQA \\
& Training samples & 765K total \\
& Coldstart epochs & 3 \\
& RL steps & 70K \\
& LR (coldstart) & $3\times10^{-5}$ \\
& LR (RL) & $1\times10^{-6}$ \\
& Batch size & 16 \\
& Per-GPU batch size & 4 \\
& Gradient accumulation & 2 \\
& Precision & bf16 \\

\midrule
\multirow{5}{*}{Hardware}
& GPUs & 4 NVIDIA H100 \\
& DeepSpeed & ZeRO-3 for 7B only \\
& Coldstart time (7B) & $\sim$3 days \\
& RL time (7B) & $\sim$3--4 days \\
& Eval hardware & same as training \\

\midrule
\multirow{6}{*}{Reward}
& Grounding reward & IoU-based accuracy \\
& IoU threshold & 0.5 \\ \cmidrule(lr){2-3}
& VQA reward & semantic answer matching \\
& VQA reward model (RL) & Qwen3-VL-4B \\
& Eval protocol & GPT-based matching \\ \cmidrule(lr){2-3}
& MCQ reward & exact option matching \\

\midrule

\multirow{1}{*}{$\mathcal{L}_{\mathrm{cold}}$} & $\lambda_{\mathrm{box}}$ & 1.0 \\

\midrule

\multirow{6}{*}{$\mathcal{L}_{\mathrm{RL}}$}
& Group size $G$ & 16 \\
& KL coefficient $\beta$ & 0.04 \\
& $\lambda_{\mathrm{fov}}$ & 1.0 \\
& $\lambda_{\mathrm{reg}}$ & 0.2 \\
& Reference model & our coldstart model trained for 3 epochs \\
& Clipping in $\mathcal{L}_{\mathrm{GRPO}\text{-}\mathrm{tok}}$ ($\epsilon$)~\cite{shao2024deepseekmath} & 0.2 \\

\bottomrule
\end{tabular}
}
\end{table}

\smallbreak\noindent\textbf{Prompt and output format.}
In all experiments, we place the question before the instruction prompt, which asks the model to generate both an intermediate reasoning process and the final answer in a structured format as follows:
\begin{tcolorbox}[
    colback=gray!10,
    colframe=black!70,
    boxrule=0.8pt,
    arc=2mm,
    left=2mm,
    right=2mm,
    top=1.5mm,
    bottom=1.5mm
]
    \small
    \{Question\} First provide the reasoning process within \texttt{<think>}\texttt{</think>} and then output the answer in \texttt{<answer>}\texttt{</answer>} tags.
\end{tcolorbox}
\noindent
Note that this prompt does not explicitly instruct the model to emit the \texttt{<fov>} token or perform foveation.
Instead, foveation behavior (\ie, emitting \texttt{<fov>}) is learned during the coldstart stage, where the same prompt format is used for supervision.
Thus, even without explicit mention of foveation in the prompt, the model is capable of invoking \texttt{<fov>} when high-res visual evidence is needed.

\smallbreak\noindent\textbf{Hardware and training setup.}
Table~\ref{tab:hyperparams} summarizes key hyperparameters, main hardware configurations, and training settings used in our experiments.

\clearpage

\begin{figure}[t]
\captionsetup{type=algorithm}
\centering
\begin{minipage}{0.96\linewidth}
\begin{minted}{python}
blocks = []
for sample in batch:
    # Generate one token from the token policy
    y_t, h_t = pi_theta(sample)

    if y_t != "<fov>":
        # Standard text-token generation
        block = [y_t]
    else:
        # Retrieve variable-length high-res visual tokens
        V_t = retrieve_highres_tokens(sample.image, pi_phi(h_t))
        block = ["<fov>"] + V_t + ["</fov>"]

    blocks.append(block)

# Match all inserted blocks to the same length in the batch
L_max = max(len(block) for block in blocks)
for sample, block in zip(batch, blocks):
    pad_len = L_max - len(block)

    # Left-pad the previous sequence, then append the new block
    sample.input_ids = [PAD] * pad_len + sample.input_ids + block

    # Left-pad tensors to keep batch shapes aligned
    sample.attention_mask = [0] * pad_len + sample.attention_mask + [1] * len(block)
    sample.hidden_states = concat(zero_tensor(pad_len), sample.hidden_states)
    sample.kv_cache = concat(zero_tensor(pad_len), sample.kv_cache)
    sample.pos_ids = [PAD_POS] * pad_len + sample.pos_ids + new_pos_ids(block)
\end{minted}
\end{minipage}
\vspace{-1.0mm}
\caption{Pseudocode for \methodabbr~batched decoding. For shape-compatible batch, we prepend zero-padding to the existing sequence and append the new block.}
\vspace{-5.0mm}
\label{code:batched_variable_decode}
\end{figure}

\subsection{Inference details}
\label{sec:supp_inference}

\smallbreak\noindent\textbf{Variable-length token generation.}
A standard autoregressive VLM generates {\em exactly one token} at each decoding step.
Our model also generates one token at a step if it is a standard text token (\ie, $y_t \in \mathcal{V}$).
However, when the model triggers foveation (\ie, $y_t = \texttt{<fov>}$), the VLM output is no longer a single token but a {\em variable number of tokens} as follows:
\begin{align}
    \texttt{<fov>},\ \mathbf{V}_{t,1},\ \mathbf{V}_{t,2},\ \ldots,\ \mathbf{V}_{t,m_t},\ \texttt{</fov>},
\end{align}
where $\mathbf{V}_t \in \mathbb{R}^{m_t \times d}$ denotes the retrieved high-res visual tokens, whose length is $m_t$.
This creates a practical {\em batching issue} during inference:
Different samples in the same batch may produce different numbers of new tokens at the same decoding step, which can no longer be ``batchified'' since batched processing requires all samples to have the same shape in the standard PyTorch implementation.

To resolve this, after each decoding step, we first identify the maximum number of new tokens among all samples in the batch.
We then zero-pad every other sample so that all samples have the same extended length.
In this way, all tensors in the batch remain shape-compatible even though the actual number of generated tokens differs across samples.
In our implementation, this padding is applied to attention mask, hidden states, key--value cache, position ids~\cite{su2021roformer} inside the transformers decoding loop.
This enables batched AR decoding even when the number of newly generated tokens varies across samples in a batch.
Algorithm.~\ref{code:batched_variable_decode} provides pseudocode for this batched-decoding procedure.

\clearpage

\smallbreak\noindent\textbf{Stopping and budget control during generation.}
At each decoding step, the token policy $\pi_\theta$ adaptively determines whether to generate (i) a standard text token in $\mathcal{V}$, (ii) \texttt{<fov>} to trigger foveation step, or (iii) \texttt{<EOS>} to terminate generation.
Note that the number of foveation steps is not fixed in advance, but is determined dynamically.
For practical stability, we enforce two limits during generation: a maximum total generation length $T_{\max}=2048$, and a maximum number of allowed foveation steps $K_{\max}=4$.
If the total generation length reaches $T_{\max}$, generation is terminated immediately by forcing $\texttt{<EOS>}$.
If the number of \texttt{<fov>} tokens reaches $K_{\max}$, we prevent any additional foveation by forcing the probability of \texttt{<fov>} to zero by setting its logit to $-\infty$ before softmax.

\setlength{\aboverulesep}{0pt}
\setlength{\belowrulesep}{0pt}

\begin{table*}[t]
\centering
\small
\setlength{\tabcolsep}{3.5pt}
\renewcommand{\arraystretch}{1.12}
\caption{Statistics of the coldstart training data. Superscripts \textsuperscript{hmn} and \textsuperscript{psd} denote the human-annotated GT boxes and our VLM-generated pseudo boxes, respectively. $W$ and $H$ denote the average image width and height, $\rho$ denotes the average fraction of image area covered by the box, and $T_{\mathrm{fov}}$ denotes the average number of foveation labels. $\Delta$ indicates the difference between pseudo and human supervision.}
\label{tab:coldstart_dataset_stats}
\resizebox{\textwidth}{!}{
\begin{tabular}{l cccc c ccc c ccc c}
\toprule
\multirow{2}{*}{Bench}
& \multicolumn{4}{c}{Doc/Text/Chart}
& \multicolumn{1}{c}{Gen.}
& \multicolumn{3}{c}{Relation}
& \multicolumn{1}{c}{Fine}
& \multicolumn{3}{c}{Grounding}
& \multirow{2}{*}{Avg} \\
\cmidrule(lr){2-5}
\cmidrule(lr){6-6}
\cmidrule(lr){7-9}
\cmidrule(lr){10-10}
\cmidrule(lr){11-13}
& DocV & TxtC & TxtV & Info
& F30k
& GQA & OI & VSR
& CUB
& RefC & RefC+ & RefCg
& \\
\midrule
W         & 1768.5 & 938.7 & 939.8 & 1201.8 & 459.3 & 493.9 & 963.1 & 577.5 & 465.6 & 595.7 & 595.5 & 584.1 & 798.6 \\
H         & 2129.5 & 826.5 & 824.8 & 2442.4 & 395.5 & 409.5 & 798.7 & 501.9 & 386.9 & 485.0 & 485.1 & 479.6 & 847.1 \\
\midrule
$\rho^{\mathrm{hmn}}$ & 5.4 & 4.3 & 4.4 & 8.8 & 14.4 & 18.0 & 21.9 & 36.8 & 35.5 & 20.9 & 21.1 & 20.6 & 17.7 \\
$\rho^{\mathrm{psd}}$ & 2.5 & 11.3 & 10.2 & 6.7 & 16.3 & 14.2 & 15.5 & 13.5 & 14.0 & 15.2 & 15.7 & 14.9 & 12.5 \\
\rowcolor{lightgraydelta}
$\Delta \rho$ & $-2.9$ & $+7.0$ & $+5.8$ & $-2.1$ & $+1.9$ & $-3.8$ & $-6.4$ & $-23.3$ & $-21.5$ & $-5.7$ & $-5.4$ & $-5.7$ & $-5.2$ \\
\midrule
$T^{\mathrm{hmn}}_{\mathrm{fov}}$ & 1.0 & 1.0 & 1.0 & 1.0 & 1.0 & 1.0 & 1.0 & 1.0 & 1.0 & 1.0 & 1.0 & 1.0 & 1.0 \\
$T^{\mathrm{psd}}_{\mathrm{fov}}$ & 2.9 & 2.0 & 1.8 & 3.7 & 1.5 & 1.6 & 1.5 & 1.0 & 1.6 & 1.1 & 1.1 & 1.2 & 1.8 \\
\rowcolor{lightgraydelta}
$\Delta T_{\mathrm{fov}}$ & $+1.9$ & $+1.0$ & $+0.8$ & $+2.7$ & $+0.5$ & $+0.6$ & $+0.5$ & $0.0$ & $+0.6$ & $+0.1$ & $+0.1$ & $+0.2$ & $+0.8$ \\
\bottomrule
\end{tabular}
}
\end{table*}

\subsection{Coldstart data statistics}
\label{sec:supp_coldstart}
Table~\ref{tab:coldstart_dataset_stats} summarizes statistics of the coldstart training data.
Comparing the average image-area ratio $\rho$ covered by the human-annotated GT boxes and our pseudo boxes, we observe that $\rho^{\mathrm{psd}}$ is lower than $\rho^{\mathrm{hmn}}$ on average (12.5 {\em vs.}\ 17.7), while $T^{\mathrm{psd}}_{\mathrm{fov}}$ is slightly higher than $T^{\mathrm{hmn}}_{\mathrm{fov}}$ (1.8 {\em vs.}\ 1.0).
This suggests that pseudo supervision encourages a more localized evidence-acquisition strategy: instead of revealing a single large (human-annotated) region, it guides the model to perform multiple targeted foveations.
Such behavior is analogous to human visual perception, where task-relevant evidence is often acquired progressively through a sequence of fixations rather than revealed all at once in a single glance.

As analyzed further in Sec.~\ref{sec:supp_abl_supervision}, we find that training with pseudo labels is more effective than training with human-annotated labels.
We hypothesize that this is because human-annotated boxes tend to {\em tightly localize the most relevant region}, whereas the VLM may instead benefit from {\em a sequential form of targeted evidence}, such as multiple nearby cues or a sequence of smaller glimpses that progressively refine its focus.
This suggests that effective foveation supervision should reflect the model's actual demands (\eg, sequential foveation to reduce uncertainty, validate intermediate predictions, or revisit key evidence), rather than relying on a single region that appears most relevant to a human annotator.

\clearpage

\begin{table*}[t]
\centering
\small
\setlength{\tabcolsep}{4pt}
\renewcommand{\arraystretch}{1.12}
\caption{Ablation study on {\em test-time} foveation count and box-parameter. Starting from the {\em fov-only} 3B coldstart model in Table~\ref{tab:fovonly} (d), we intervene by restricting the maximum number of generated foveation actions $T_{\mathrm{fov}}^{\mathrm{max}}$ and by overriding either the box center $(c^x,c^y)$ or size $(w,h)$ during inference time. Accuracy on DocVQA~\cite{mathew2021docvqa}, TextVQA~\cite{singh2019textvqa}, and Visual7W~\cite{zhu2016visual7w} is evaluated using Qwen3-4B-based~\cite{yang2025qwen3} semantic string matching.}
\label{tab:ablation_box_params}
\resizebox{\textwidth}{!}{
\begin{tabular}{c c cc cccc cccc cccc}
\toprule
\multirow{2}{*}{Model} &
\multirow{2}{*}{$T_{\mathrm{fov}}^{\mathrm{max}}$} &
\multicolumn{2}{c}{box action $b$} &
\multicolumn{4}{c}{DocV~\cite{mathew2021docvqa} (\textbf{1787$\times$2101})} &
\multicolumn{4}{c}{TxtV~\cite{singh2019textvqa} (\textbf{941$\times$820})} &
\multicolumn{4}{c}{V7W~\cite{zhu2016visual7w} (\textbf{556$\times$455})} \\
\cmidrule(lr){3-4}
\cmidrule(lr){5-8}
\cmidrule(lr){9-12}
\cmidrule(lr){13-16}
 & & $(c^x,c^y)$ & $(w,h)$
 & $N_{\mathrm{fov}}$ & $\rho$ & $T_{\mathrm{fov}}$ & acc
 & $N_{\mathrm{fov}}$ & $\rho$ & $T_{\mathrm{fov}}$ & acc
 & $N_{\mathrm{fov}}$ & $\rho$ & $T_{\mathrm{fov}}$ & acc \\
\midrule
(a) ours$^{\text{3B}}_{\text{cold}}$ & 3 & learned & learned
& 362.9 & 25.3 & 1.9 & 75.7
& 242.2 & 20.8 & 1.4 & 85.7
& 72.1 & 17.2 & 1.1 & 56.7 \\
\midrule
(b) & 1 & learned & learned
& 209.3 & 20.7 & 1 & 72.5
& 187.4 & 19.3 & 1 & 86.4
& 56.9 & 16.0 & 0.9 & 57.2 \\
(c) & 1 & center & learned
& 227.2 & 22.4 & 1 & 54.5
& 199.1 & 20.5 & 1 & 79.4
& 59.3 & 16.5 & 0.9 & 56.0 \\
(d) & 1 & random $(-1,1)$ & learned
& 171.0 & 16.9 & 1 & 52.3
& 153.1 & 15.8 & 1 & 76.9
& 45.0 & 12.3 & 0.9 & 56.7 \\
(e) & 1 & learned & fixed $(.5,.5)$
& 62.2 & 6.1 & 1 & 57.1
& 58.5 & 6.0 & 1 & 82.8
& 19.2 & 5.3 & 0.9 & 56.9 \\
(f) & 1 & learned & random $(0,1)$
& 57.8 & 5.7 & 1 & 54.9
& 56.7 & 5.9 & 1 & 79.2
& 19.1 & 5.4 & 0.9 & 57.3 \\
(g) & 1 & random $(-1,1)$ & random $(0,1)$
& 50.1 & 4.9 & 1 & 43.3
& 48.2 & 4.9 & 1 & 75.9
& 13.9 & 3.9 & 0.9 & 56.0 \\

\midrule

Baseline~\cite{bai2025qwen25vl} & - & - & -
& - & - & - & 55.1
& - & - & - & 77.2
& - & - & - & 50.5 \\
\bottomrule
\end{tabular}
}
\vspace{-3.0mm}
\end{table*}

\section{Additional Experimental Analyses}
\label{sec:supp_results}
\vspace{-2.0mm}

\subsection{Test-time ablation on foveation count and box action}
\label{sec:supp_abl_foveation}

To investigate the effect of our learned foveations, we perform a set of {\em test-time} interventions on our model.
Specifically, during decoding, we (i) restrict the maximum number of generated foveation actions $T_{\mathrm{fov}}^{\mathrm{max}}$, and (ii) override the predicted box center $(c^x,c^y)$ or size $(w,h)$\footnote{The center $(c^x,c^y)$ is normalized to $[-1,1]$, where $(-1,-1)$ and $(1,1)$ correspond to the top-left and bottom-right of the image, respectively. The size $(w,h)$ is normalized to $[0,1]$; the box width (or height) of 1 spans 50\% of the image width (or height).} with random (or fixed) values, allowing us to separately examine the importance of the foveation budget and box location \& scale without retraining.
Table~\ref{tab:ablation_box_params} summarizes the results.

\smallbreak\noindent\textbf{Effect of limiting the foveation count.}
Comparing (a) and (b), enforcing at most ``one'' foveation causes a noticeable drop on DocVQA (75.7 $\rightarrow$ 72.5), while the change is relatively minor on TextVQA and Visual7W.
This suggests that multiple foveations are occasionally helpful for visually demanding document-style inputs (\eg, DocVQA), where the answer may require sequential evidence acquisition from multiple distant regions (Fig.~\ref{fig:supp_qual2} (c)).
On TextVQA and Visual7W, the original model (a) already uses only a modest $T_{\text{fov}}^{\text{max}}$ on average (1.1-1.4), thus forcing $T_{\text{fov}}^{\text{max}}$ down to 1.0 does not drastically alter its behavior.

\smallbreak\noindent\textbf{Importance of learning {\em where} to foveate.}
Comparing (b) against (c) and (d), replacing the learned box center with either a fixed image center (c) or a random location (d) substantially harms accuracy on DocVQA and TextVQA, while having little effect on Visual7W.
This suggests that, for document and text-centric images, success primarily depends on locating task-relevant evidence that may appear anywhere in the image (\eg, in corners or small peripheral regions).
In contrast, on Visual7W, forcing the box center to be either centered or random leads to only a minor performance drop.
We attribute this to the nature of Visual7W images, where the task-relevant object is often already dominantly visible and frequently located near the image center as illustrated in Fig.~\ref{fig:supp_qua1}.

\smallbreak\noindent\textbf{Importance of learning \emph{how much} to foveate.}
Comparing (b) against (e) and (f), overriding the box size with either a fixed $(0.5,0.5)$ box or a random size also leads to clear degradation on DocVQA and TextVQA, whereas the impact remains small on Visual7W.
The results suggest that the scale of the required evidence is highly instance-dependent in text-heavy benchmarks:
some questions require a very small local crop, while others benefit from a wider region containing neighboring context.
A fixed or random box size therefore may fail to match the size of the truly relevant region.
Visual7W images, however, mostly contain a relatively salient object already visible under low resolution (Fig.~\ref{fig:supp_qua1}), thus making precise scale adaptation less critical.

\smallbreak\noindent\textbf{Joint corruption of center and size.}
Model (g), which randomizes both center and size of the box, yields the worst performance overall, confirming the importance of both components;
the model must learn not only ``where to look'', but also ``how much to look at''.
Again, the drop on Visual7W is small since the low-res input already contains enough information for a reasonable prediction, making the model less dependent on fine-grained evidence.

\smallbreak\noindent\textbf{Random foveation can be worse than no foveation.}
Comparing the baseline with (g), we observe that on DocVQA and TextVQA, fully random foveation is substantially worse than using no foveation at all.
A plausible explanation is that randomly retrieved high-res tokens introduce irrelevant and potentially distracting visual details, which interfere with answer generation rather than helping it.
In other words, fine-grained evidence acquisition is beneficial {\em only when it is targeted}.
On Visual7W, however, (g) still surpasses the baseline (56.0 vs.\ 50.5), likely because many Visual7W images contain a large, dominant object, so even random foveation can still capture useful parts of it.

\smallbreak\noindent\textbf{A single foveation is already highly effective.}
Finally, comparing the baseline with (b) shows that even a single foveation provides a large gain across all three benchmarks: 55.1 $\rightarrow$ 72.5 on DocVQA, 77.2 $\rightarrow$ 86.4 on TextVQA, and 50.5 $\rightarrow$ 57.2 on Visual7W.
The results show that the gain comes not simply from observing more visual tokens, but from retrieving the {\em right} high-resolution region.
For example, although (c) consumes more visual tokens than (b), it performs much worse, indicating that additional visual input is not helpful unless it captures task-relevant evidence.
Consistently, when the box is poorly chosen, as in (c)--(g), accuracy drops sharply.
This shows that a single foveation can be highly effective only when guided by a learned policy: $\pi_{\phi}$.

\smallbreak\noindent\textbf{Takeaway.}
Overall, these experiments show that the benefit of learned foveation does not come from na\"ively revealing more high-res pixels, but from acquiring the {\em right evidence} with the {\em right scale}.
This benefit is especially apparent on scenarios where task-relevant cues are sparse and spatially dispersed (\eg, DocV~\cite{mathew2021docvqa} and TxtV~\cite{singh2019textvqa}).
Taken together, the results support our central view of foveation as ``an active evidence-acquisition mechanism'', rather than a simple token-increase strategy.
Such behavior is also consistent with human visual perception, where effective perception depends not on seeing more everywhere, but on selectively attending to the most informative region for the target task.

\clearpage

\setlength{\aboverulesep}{0.5pt}
\setlength{\belowrulesep}{0.5pt}

\begin{table*}[t]
\centering
\small
\setlength{\tabcolsep}{4pt}
\renewcommand{\arraystretch}{1.12}
\caption{Comparison between two types of foveation supervision: human-annotated GT boxes (ours$^{\mathrm{3B}}_{\mathrm{hmn}}$) and VLM-generated pseudo boxes (ours$^{\mathrm{3B}}_{\mathrm{psd}}$). Accuracy is evaluated using Qwen3-4B-based semantic matching. $T_{\mathrm{fov}}$ denotes the number of foveations. $N_{\mathrm{fov}}$ counts all retrieved visual tokens via foveations, {\em including those from overlapping foveation regions}, whereas $\rho$ measures only the total {\em unique high-res image area} revealed to the model, ignoring overlap. $\Delta$ denotes the difference for each metric.}
\vspace{3.0mm}
\label{tab:cold_gt_vs_pseudo}
\resizebox{0.95\textwidth}{!}{
\begin{tabular}{l c c c c c c c c c c c c}
\toprule
\multirow{2}{*}{Bench}
& \multicolumn{3}{c}{Accuracy}
& \multicolumn{3}{c}{$T_{\mathrm{fov}}$}
& \multicolumn{3}{c}{$\rho$}
& \multicolumn{3}{c}{$N_{\mathrm{fov}}$}
\rule[-1.3ex]{0pt}{4.0ex}\\

\cmidrule(lr){2-4}
\cmidrule(lr){5-7}
\cmidrule(lr){8-10}
\cmidrule(lr){11-13}
& ours$^{\mathrm{3B}}_{\mathrm{hmn}}$ & ours$^{\mathrm{3B}}_{\mathrm{psd}}$ & \multicolumn{1}{>{\columncolor{lightgraydelta}}c}{$\Delta$}
& ours$^{\mathrm{3B}}_{\mathrm{hmn}}$ & ours$^{\mathrm{3B}}_{\mathrm{psd}}$ & \multicolumn{1}{>{\columncolor{lightgraydelta}}c}{$\Delta$}
& ours$^{\mathrm{3B}}_{\mathrm{hmn}}$ & ours$^{\mathrm{3B}}_{\mathrm{psd}}$ & \multicolumn{1}{>{\columncolor{lightgraydelta}}c}{$\Delta$}
& ours$^{\mathrm{3B}}_{\mathrm{hmn}}$ & ours$^{\mathrm{3B}}_{\mathrm{psd}}$ & \multicolumn{1}{>{\columncolor{lightgraydelta}}c}{$\Delta$}
\rule[-1.3ex]{0pt}{4.0ex}\\

\midrule
DocV & 44.5 & 52.6 & \multicolumn{1}{>{\columncolor{lightgraydelta}}c}{\textcolor{darkgreenacc}{$+8.1$}} & 1.0 & 2.2 & \multicolumn{1}{>{\columncolor{lightgraydelta}}c}{$+1.2$} & 3.9 & 2.8 & \multicolumn{1}{>{\columncolor{lightgraydelta}}c}{$-1.1$} & 39.3 & 35.9 & \multicolumn{1}{>{\columncolor{lightgraydelta}}c}{$-3.4$} \\

TxtC & 68.0 & 72.4 & \multicolumn{1}{>{\columncolor{lightgraydelta}}c}{\textcolor{darkgreenacc}{$+4.4$}} & 1.0 & 1.5 & \multicolumn{1}{>{\columncolor{lightgraydelta}}c}{$+0.5$} & 3.2 & 7.4 & \multicolumn{1}{>{\columncolor{lightgraydelta}}c}{$+4.2$} & 31.7 & 84.1 & \multicolumn{1}{>{\columncolor{lightgraydelta}}c}{$+52.4$} \\

TxtV & 74.6 & 81.2 & \multicolumn{1}{>{\columncolor{lightgraydelta}}c}{\textcolor{darkgreenacc}{$+6.6$}} & 1.0 & 1.4 & \multicolumn{1}{>{\columncolor{lightgraydelta}}c}{$+0.4$} & 3.5 & 7.1 & \multicolumn{1}{>{\columncolor{lightgraydelta}}c}{$+3.6$} & 33.5 & 78.1 & \multicolumn{1}{>{\columncolor{lightgraydelta}}c}{$+44.6$} \\

Dude & 34.0 & 39.1 & \multicolumn{1}{>{\columncolor{lightgraydelta}}c}{\textcolor{darkgreenacc}{$+5.1$}} & 1.0 & 2.1 & \multicolumn{1}{>{\columncolor{lightgraydelta}}c}{$+1.1$} & 4.5 & 3.6 & \multicolumn{1}{>{\columncolor{lightgraydelta}}c}{$-0.9$} & 44.4 & 45.0 & \multicolumn{1}{>{\columncolor{lightgraydelta}}c}{$+0.6$} \\

Sroi & 49.1 & 56.7 & \multicolumn{1}{>{\columncolor{lightgraydelta}}c}{\textcolor{darkgreenacc}{$+7.6$}} & 1.0 & 2.3 & \multicolumn{1}{>{\columncolor{lightgraydelta}}c}{$+1.3$} & 2.9 & 2.9 & \multicolumn{1}{>{\columncolor{lightgraydelta}}c}{$0.0$} & 19.3 & 26.3 & \multicolumn{1}{>{\columncolor{lightgraydelta}}c}{$+7.0$} \\

Info & 37.0 & 36.1 & \multicolumn{1}{>{\columncolor{lightgraydelta}}c}{\textcolor{darkredacc}{$-0.9$}} & 1.0 & 2.6 & \multicolumn{1}{>{\columncolor{lightgraydelta}}c}{$+1.6$} & 3.9 & 4.4 & \multicolumn{1}{>{\columncolor{lightgraydelta}}c}{$+0.5$} & 26.5 & 39.9 & \multicolumn{1}{>{\columncolor{lightgraydelta}}c}{$+13.4$} \\
\midrule

F30k & 61.8 & 62.2 & \multicolumn{1}{>{\columncolor{lightgraydelta}}c}{\textcolor{darkgreenacc}{$+0.4$}} & 1.0 & 1.2 & \multicolumn{1}{>{\columncolor{lightgraydelta}}c}{$+0.2$} & 11.5 & 11.5 & \multicolumn{1}{>{\columncolor{lightgraydelta}}c}{$0.0$} & 26.2 & 29.7 & \multicolumn{1}{>{\columncolor{lightgraydelta}}c}{$+3.5$} \\

V7W & 56.8 & 56.7 & \multicolumn{1}{>{\columncolor{lightgraydelta}}c}{\textcolor{darkredacc}{$-0.1$}} & 1.0 & 1.1 & \multicolumn{1}{>{\columncolor{lightgraydelta}}c}{$+0.1$} & 10.9 & 9.3 & \multicolumn{1}{>{\columncolor{lightgraydelta}}c}{$-1.6$} & 35.9 & 35.5 & \multicolumn{1}{>{\columncolor{lightgraydelta}}c}{$-0.4$} \\
\midrule

GQA & 61.2 & 62.9 & \multicolumn{1}{>{\columncolor{lightgraydelta}}c}{\textcolor{darkgreenacc}{$+1.7$}} & 1.0 & 1.3 & \multicolumn{1}{>{\columncolor{lightgraydelta}}c}{$+0.3$} & 10.4 & 9.8 & \multicolumn{1}{>{\columncolor{lightgraydelta}}c}{$-0.6$} & 26.4 & 28.7 & \multicolumn{1}{>{\columncolor{lightgraydelta}}c}{$+2.3$} \\

OI  & 81.5 & 82.3 & \multicolumn{1}{>{\columncolor{lightgraydelta}}c}{\textcolor{darkgreenacc}{$+0.8$}} & 1.0 & 1.1 & \multicolumn{1}{>{\columncolor{lightgraydelta}}c}{$+0.1$} & 12.6 & 10.5 & \multicolumn{1}{>{\columncolor{lightgraydelta}}c}{$-2.1$} & 121.6 & 113.0 & \multicolumn{1}{>{\columncolor{lightgraydelta}}c}{$-8.6$} \\

VSR & 63.4 & 65.4 & \multicolumn{1}{>{\columncolor{lightgraydelta}}c}{\textcolor{darkgreenacc}{$+2.0$}} & 1.0 & 0.8 & \multicolumn{1}{>{\columncolor{lightgraydelta}}c}{$-0.2$} & 17.8 & 8.8 & \multicolumn{1}{>{\columncolor{lightgraydelta}}c}{$-9.0$} & 64.4 & 33.5 & \multicolumn{1}{>{\columncolor{lightgraydelta}}c}{$-30.9$} \\
\midrule

CUB & 83.7 & 80.1 & \multicolumn{1}{>{\columncolor{lightgraydelta}}c}{\textcolor{darkredacc}{$-3.6$}} & 1.0 & 1.2 & \multicolumn{1}{>{\columncolor{lightgraydelta}}c}{$+0.2$} & 21.6 & 11.0 & \multicolumn{1}{>{\columncolor{lightgraydelta}}c}{$-10.6$} & 49.0 & 25.5 & \multicolumn{1}{>{\columncolor{lightgraydelta}}c}{$-23.5$} \\

\midrule
Avg & 59.6 & 62.3 & \multicolumn{1}{>{\columncolor{lightgraydelta}}c}{\textcolor{darkgreenacc}{$+2.7$}} & 1.0 & 1.6 & \multicolumn{1}{>{\columncolor{lightgraydelta}}c}{$+0.6$} & 8.9 & 7.4 & \multicolumn{1}{>{\columncolor{lightgraydelta}}c}{$-1.5$} & 43.2 & 47.9 & \multicolumn{1}{>{\columncolor{lightgraydelta}}c}{$+4.7$} \\

\bottomrule
\end{tabular}
}
\end{table*}

\subsection{Pseudo foveation supervision vs.\ human-annotated GT boxes}
\vspace{-2.0mm}
\label{sec:supp_abl_supervision}
To compare the effect of different supervision signals for the foveation policy $\pi_{\phi}$, we train two coldstart 3B models.
The first model (ours$^{\mathrm{3B}}_{\mathrm{hmn}}$) uses human-annotated GT boxes.
The second model (ours$^{\mathrm{3B}}_{\mathrm{psd}}$) uses VLM-generated pseudo box labels (cf. Sec.~\ref{sec:training}).\footnote{In this experiment, both models are trained only on datasets with GT boxes (\eg, RefCOCO and Visual CoT); ScienceQA is excluded due to the absence of GT boxes.}
Table~\ref{tab:cold_gt_vs_pseudo} summarizes the results.

Even without human supervision, ours$^{\mathrm{3B}}_{\mathrm{psd}}$ yields higher average accuracy than ours$^{\mathrm{3B}}_{\mathrm{hmn}}$ (62.3 {\em vs.}\ 59.6) with a slightly larger number of foveations on average (1.6 {\em vs.}\ 1.0).
Interestingly, ours$^{\mathrm{3B}}_{\mathrm{psd}}$ improves accuracy while {\em reducing} $\rho$ (7.4 {\em vs.}\ 8.9)---\ie, the model observes a smaller fraction of the original high-res image yet makes better predictions.
We attribute this to the following misalignment: human-annotated boxes tend to {\em tightly localize} the region that appears most relevant to a human annotator, whereas the VLM may benefit from a different evidence-acquisition pattern, such as multiple distant glimpses, broader surrounding context, or a sequence of progressively refined crops.
We also observe that ours$^{\mathrm{3B}}_{\mathrm{psd}}$ sees more visual tokens ($\Delta N_{\mathrm{fov}}=+4.7$) while revealing a smaller high-res image area ($\Delta \rho=-1.5$).
The results imply that the model requests multiple foveations to ``revisit'' previously observed regions, resulting in repeated (or slightly shifted) glimpses of the same local region.
Such behavior can be beneficial when the model needs to verify the evidence by (i) refining uncertain evidence or (ii) exploring nearby context at different spatial extents.

Overall, these results suggest that supervising foveation with a single (human-annotated) box may be suboptimal.
What matters is not just seeing one salient region, but learning {\em how to gather the evidence} that the model actually needs through {\em test-time refinement and exploration}.

\clearpage

\begin{figure}[t]
    \centering
    \scalebox{0.35}{
    \includegraphics{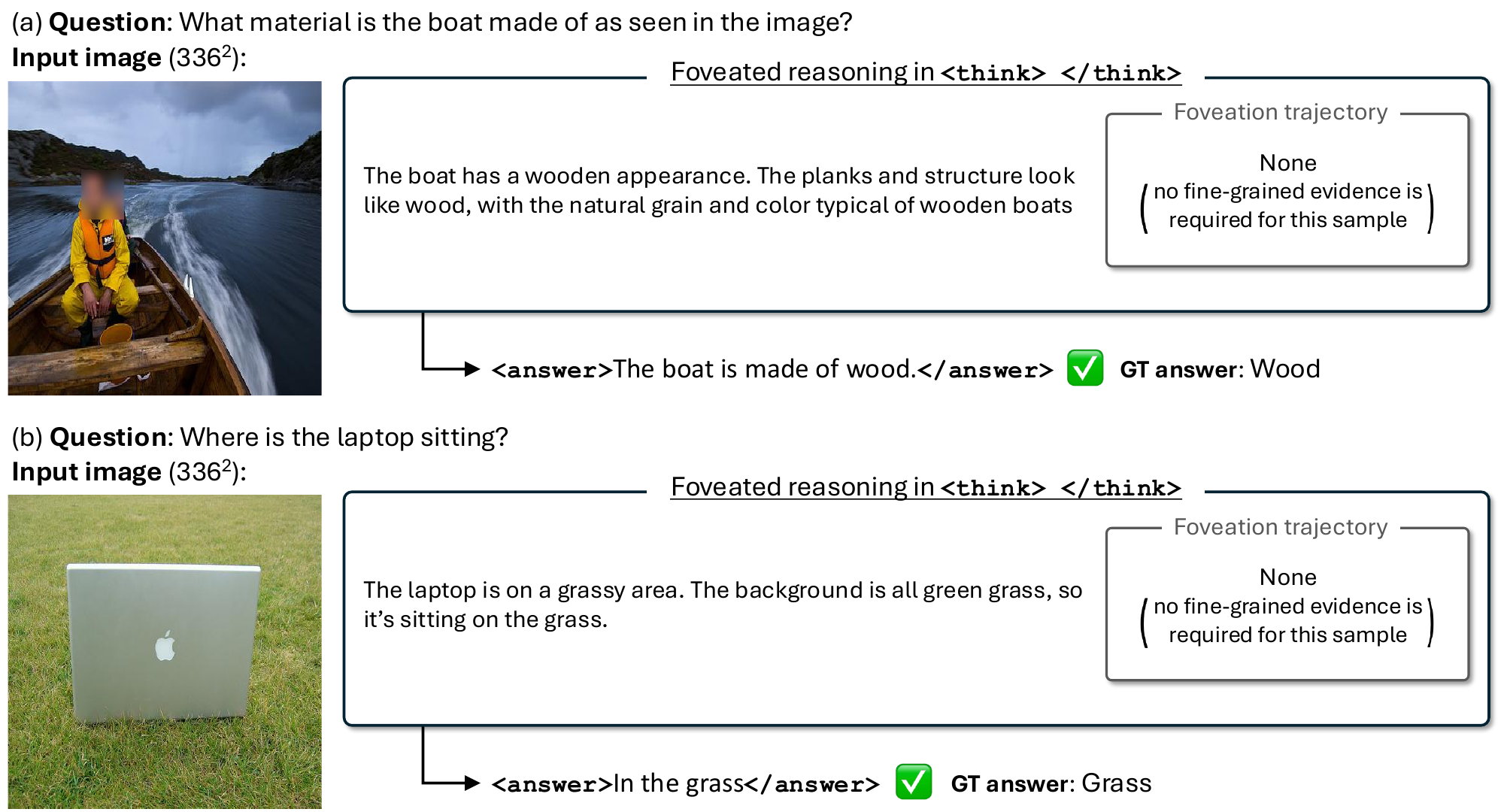}}
    \caption{Cases where \methodabbr\ answers correctly without foveation~\cite{plummer2016flickr30k, zhu2016visual7w}.}
    \label{fig:supp_qua1}
    \vspace{-5.0mm}
\end{figure}

\subsection{Qualitative results and analyses}
\label{sec:supp_qual}

Figures~\ref{fig:supp_qua1} and~\ref{fig:supp_qual2} provide qualitative examples of how \methodabbr\ uses foveation in a task-adaptive manner.
Figure~\ref{fig:supp_qua1} shows cases where the model predicts correct answer {\em without triggering any foveation};
the task-relevant evidence is already sufficiently salient in the low-res input, suggesting that the model has learned {\em not to invoke foveation unnecessarily} when it is given sufficient evidence already.
In other words, the model uses no foveation by default, but only when additional evidence is needed.

In Fig.~\ref{fig:supp_qual2} (a), the question requires checking two small, distant attributes (wing and eye colors) so the model performs {\em two separate} foveations, one for each localized cue.
In Fig.~\ref{fig:supp_qual2} (b), the question also asks about two attributes (bill shape and forehead color) but both cues are spatially close, so a {\em single} foveation suffices.
This indicates that the model foveates efficiently in the sense that it increases the number of foveations only when the evidence is distributed across distant regions, while avoiding unnecessary foveations when multiple evidence can be verified with a single foveation.

In Fig.~\ref{fig:supp_qual2} (c), the first foveation already reveals the main clue needed for answering the question, namely the map title.
However, the model continues to explore additional regions, including the legend and scale bar, before producing the final answer.
This indicates that \methodabbr\ does not always stop once a plausible answer becomes available; when the initial evidence is still insufficient (or ambiguous), the model may sacrifice efficiency in favor of confidence, gathering additional evidence before committing to a final prediction.

These examples reveal that \methodabbr\ has learned a flexible trade-off between efficiency and reliability.
When the answer is already evident from the coarse input, it avoids foveation altogether.
When fine-grained evidence is required, it uses as few foveations as possible; yet under uncertainty, it is also willing to perform exploratory foveations to improve prediction reliability.

\begin{figure}[t]
    \centering
    \scalebox{0.35}{
    \includegraphics{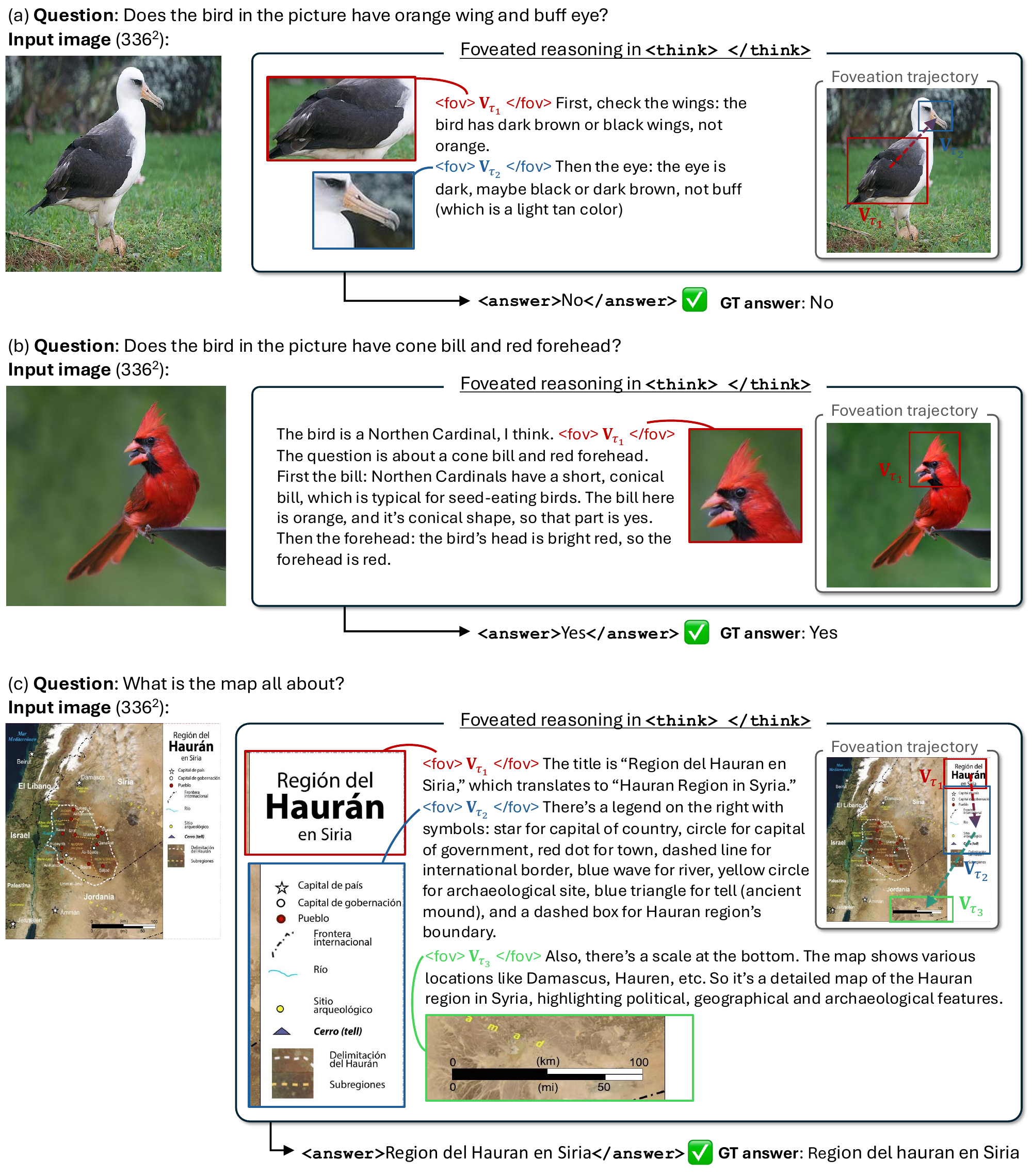}}
    \caption{Qualitative examples of task-adaptive foveation in \methodabbr. (a,b) Examples from CUB~\cite{wah2011cub}: in (a), the model performs two foveations because the queried attributes are small and spatially distant, whereas in (b), a single foveation suffices since the relevant attributes are spatially close. (c) An example from DUDE~\cite{vanlandeghem2023dude}: although the first foveation already reveals an important clue, the model performs additional exploratory foveations to reduce ambiguity, thereby making a more reliable prediction.}
    \label{fig:supp_qual2}
    \vspace{-5.0mm}
\end{figure}





\clearpage

\section{Limitations and Future Work}
\label{sec:supp_futurework}
Despite its effectiveness, our current framework still poses several limitations, suggesting important directions for future work.

First, each newly acquired high-res evidence is \emph{accumulated} in the Transformer key--value cache (Eq.~\eqref{eq:cache_growing}).
While this design preserves hidden-state continuity across foveation steps, it causes a linear growth in memory \& compute with the amount of acquired evidence, which may limit scalability in longer-horizon settings.
A natural next step is to replace this cache-growing mechanism with a more compact architecture, such as a state-space model~\cite{gu2024mamba,fang2025ahn}, which could scale more gracefully with the number of foveation steps while retaining the stateful integration of newly acquired evidence.

Second, our empirical validation is currently focused on single-image benchmarks, whereas the proposed mechanism could be applicable to temporally extended settings: video understanding.
Compared with static images, video contains much stronger pixel redundancy across adjacent frames, making dense per-frame processing highly inefficient.
In this setting, foveation offers a promising mechanism for efficient video understanding:
rather than processing all frames at a uniform resolution, the model could leverage contextual information from previous low-res frames to decide where to foveate in the current frame for high-res evidence acquisition.
This allows the model to use previously observed frames as temporal context, so that foveation in the current frame can focus on newly appeared and task-critical regions, similar to human visual attention.

Lastly, the current framework uses a relatively simple foveation policy $\pi_{\phi}$ and action space $b_t = [c^x_t, c^y_t, w_t, h_t]$.
While this simple design is sufficient to demonstrate the efficacy of stateful action-based evidence acquisition, richer policies may further improve efficiency and robustness.
For example, future work could allow multiple-region acquisition when relevant evidence is spatially distributed, or adaptive-resolution foveation when the model requires different levels of visual detail.
In video domains, acquiring visual evidence at every frame would quickly become prohibitively expensive; this may require additional policies that decide which evidence to retain, update, or discard over time.
We leave this to future work.

Taken together, this work suggests a promising direction for future work: small AI agents operating under strict on-device constraints.
In many real-world applications (especially on edge devices) compute, memory, latency, and energy budgets are far more limited than in cloud environments, making it impractical to process all visual inputs uniformly at high resolution.
From this perspective, stateful and action-based evidence acquisition becomes particularly attractive, as an agent can selectively spend its limited budget only on information necessary for the current task.
Looking ahead, we expect future AI systems to operate in ``hybrid edge--cloud settings'', where {\em lightweight on-device agents} handle latency-sensitive or privacy-sensitive perception, while {\em larger cloud agents} provide heavier reasoning or broader world knowledge only when needed.
In such a regime, efficient visual focusing may serve as a key mechanism for the edge-device agents, allowing them to remain responsive and resource-aware, while still effectively interacting with more powerful external agents.

\clearpage

%
%
\bibliographystyle{splncs04}
\bibliography{main}

\begin{thebibliography}{10}
\providecommand{\url}[1]{\texttt{#1}}
\providecommand{\urlprefix}{URL }
\providecommand{\doi}[1]{https://doi.org/#1}

\bibitem{alayrac2022flamingo}
Alayrac, J.B., Donahue, J., Luc, P., Miech, A., Barr, I., Hasson, Y., Lenc, K., Mensch, A., Millican, K., Reynolds, M., Ring, R., Rutherford, E., Cabi, S., Han, T., Gong, Z., Samangooei, S., Monteiro, M., Menick, J., Borgeaud, S., Brock, A., Nematzadeh, A., Sharifzadeh, S., Binkowski, M., Barreira, R., Vinyals, O., Zisserman, A., Simonyan, K.: Flamingo: a visual language model for few-shot learning. arXiv preprint arXiv:2204.14198  (2022)

\bibitem{bai2025qwen25vl}
Bai, S., Chen, K., Liu, X., Wang, J., Ge, W., Song, S., Dang, K., Wang, P., Wang, S., Tang, J., Zhong, H., Zhu, Y., Yang, M., Li, Z., Wan, J., Wang, P., Ding, W., Fu, Z., Xu, Y., Ye, J., Zhang, X., Xie, T., Cheng, Z., Zhang, H., Yang, Z., Xu, H., Lin, J.: Qwen2.5-vl technical report. arXiv preprint arXiv:2502.13923  (2025)

\bibitem{bolya2023tokenmerging}
Bolya, D., Fu, C.Y., Dai, X., Zhang, P., Feichtenhofer, C., Hoffman, J.: Token merging: Your vit but faster. In: International Conference on Learning Representations (ICLR) (2023)

\bibitem{carvalho2025cropvlm}
Carvalho, M., Dias, H., Martins, B.: Cropvlm: Learning to zoom for fine-grained vision-language perception. arXiv preprint arXiv:2511.19820  (2025)

\bibitem{dai2023instructblip}
Dai, W., Li, J., Li, D., Tiong, A.M.H., Zhao, J., Wang, W., Li, B., Fung, P., Hoi, S.: Instructblip: Towards general-purpose vision-language models with instruction tuning. arXiv preprint arXiv:2305.06500  (2023)

\bibitem{fan2025grit}
Fan, Y., He, X., Yang, D., Zheng, K., Kuo, C.C., Zheng, Y., Narayanaraju, S.J., Guan, X., Wang, X.E.: Grit: Teaching mllms to think with images. In: Advances in Neural Information Processing Systems (NeurIPS) (2025)

\bibitem{fang2025ahn}
Fang, Y., Yu, W., Zhong, S., Ye, Q., Xiong, X., Wei, L.: Artificial hippocampus networks for efficient long-context modeling. arXiv preprint arXiv:2510.07318  (2025)

\bibitem{goyal2020powerbert}
Goyal, S., Choudhury, A.R., Raje, S.M., Chakaravarthy, V.T., Sabharwal, Y., Verma, A.: Power-bert: Accelerating bert inference via progressive word-vector elimination. In: Proc. International Conference on Machine Learning (ICML) (2020)

\bibitem{gu2024mamba}
Gu, A., Dao, T.: Mamba: Linear-time sequence modeling with selective state spaces. arXiv preprint arXiv:2312.00752  (2024)

\bibitem{guo2025deepseekr1}
Guo, D., Yang, D., Zhang, H., Song, J., Wang, P., Zhu, Q., Xu, R., Zhang, R., Ma, S., Bi, X., Zhang, X., Yu, X., Wu, Y., Wu, Z.F., Gou, Z., Shao, Z., Li, Z., Gao, Z., Liu, A., Xue, B., Wang, B., Wu, B., Feng, B., Lu, C., Zhao, C., Deng, C., Ruan, C., Dai, D., Chen, D., Ji, D., Li, E., Lin, F., Dai, F., Luo, F., Hao, G., Chen, G., Li, G., Zhang, H., Xu, H., Ding, H., Gao, H., Qu, H., Li, H., Guo, J., Li, J., Chen, J., Yuan, J., Tu, J., Qiu, J., Li, J., Cai, J.L., Ni, J., Liang, J., Chen, J., Dong, K., Hu, K., You, K., Gao, K., Guan, K., Huang, K., Yu, K., Wang, L., Zhang, L., Zhao, L., Wang, L., Zhang, L., Xu, L., Xia, L., Zhang, M., Zhang, M., Tang, M., Zhou, M., Li, M., Wang, M., Li, M., Tian, N., Huang, P., Zhang, P., Wang, Q., Chen, Q., Du, Q., Ge, R., Zhang, R., Pan, R., Wang, R., Chen, R.J., Jin, R.L., Chen, R., Lu, S., Zhou, S., Chen, S., Ye, S., Wang, S., Yu, S., Zhou, S., Pan, S., Li, S.S., Zhou, S., Wu, S., Yun, T., Pei, T., Sun, T., Wang, T., Zeng, W., Liu, W., Liang, W., Gao, W., Yu, W., Zhang, W.,
  Xiao, W.L., An, W., Liu, X., Wang, X., Chen, X., Nie, X., Cheng, X., Liu, X., Xie, X., Liu, X., Yang, X., Li, X., Su, X., Lin, X., Li, X.Q., Jin, X., Shen, X., Chen, X., Sun, X., Wang, X., Song, X., Zhou, X., Wang, X., Shan, X., Li, Y.K., Wang, Y.Q., Wei, Y.X., Zhang, Y., Xu, Y., Li, Y., Zhao, Y., Sun, Y., Wang, Y., Yu, Y., Zhang, Y., Shi, Y., Xiong, Y., He, Y., Piao, Y., Wang, Y., Tan, Y., Ma, Y., Liu, Y., Guo, Y., Ou, Y., Wang, Y., Gong, Y., Zou, Y., He, Y., Xiong, Y., Luo, Y., You, Y., Liu, Y., Zhou, Y., Zhu, Y.X., Huang, Y., Li, Y., Zheng, Y., Zhu, Y., Ma, Y., Tang, Y., Zha, Y., Yan, Y., Ren, Z.Z., Ren, Z., Sha, Z., Fu, Z., Xu, Z., Xie, Z., Zhang, Z., Hao, Z., Ma, Z., Yan, Z., Wu, Z., Gu, Z., Zhu, Z., Liu, Z., Li, Z., Xie, Z., Song, Z., Pan, Z., Huang, Z., Xu, Z., Zhang, Z., Zhang, Z.: Deepseek-r1 incentivizes reasoning in llms through reinforcement learning. Nature  \textbf{645}(8081),  633–638 (Sep 2025). \doi{10.1038/s41586-025-09422-z}, \url{http://dx.doi.org/10.1038/s41586-025-09422-z}

\bibitem{huang2025lavcot}
Huang, J., Tan, Z., Gong, S., Zeng, F., Zhou, J.T., Miao, C., Tan, H., Yao, W., Li, J.: Lav-cot: Language-aware visual cot with multi-aspect reward optimization for real-world multilingual vqa. arXiv preprint arXiv:2509.10026  (2025)

\bibitem{huang2023languageisnot}
Huang, S., Dong, L., Wang, W., Hao, Y., Singhal, S., Ma, S., Lv, T., Cui, L., Mohammed, O.K., Patra, B., Liu, Q., Aggarwal, K., Chi, Z., Bjorck, J., Chaudhary, V., Som, S., Song, X., Wei, F.: Language is not all you need: Aligning perception with language models. In: Advances in Neural Information Processing Systems (NeurIPS) (2023)

\bibitem{huang2019sroie}
Huang, Z., Chen, K., He, J., Bai, X., Karatzas, D., Lu, S., Jawahar, C.V.: Icdar2019 competition on scanned receipt ocr and information extraction. In: 2019 International Conference on Document Analysis and Recognition (ICDAR). IEEE (2019)

\bibitem{hudson2019gqa}
Hudson, D.A., Manning, C.D.: Gqa: A new dataset for real-world visual reasoning and compositional question answering. In: Proc. IEEE Conference on Computer Vision and Pattern Recognition (CVPR) (2019)

\bibitem{jiang2025tokenefficient}
Jiang, Y., Gu, J., Xue, T., Cheung, K.C., Molchanov, P., Yin, H., Liu, S.: Token-efficient vlm: High-resolution image understanding via dynamic region proposal. In: Proc. IEEE International Conference on Computer Vision (ICCV). pp. 24147--24158 (October 2025)

\bibitem{jonnalagadda2022foveater}
Jonnalagadda, A., Wang, W.Y., Manjunath, B.S., Eckstein, M.P.: Foveater: Foveated transformer for image classification. arXiv preprint arXiv:2105.14173  (2022)

\bibitem{kazemzadeh2014referitgame}
Kazemzadeh, S., Ordonez, V., Matten, M., Berg, T.L.: Referit game: Referring to objects in photographs of natural scenes. In: Conference on Empirical Methods in Natural Language Processing (EMNLP) (2014)

\bibitem{kong2022spvit}
Kong, Z., Dong, P., Ma, X., Meng, X., Sun, M., Niu, W., Shen, X., Yuan, G., Ren, B., Qin, M., Tang, H., Wang, Y.: Spvit: Enabling faster vision transformers via soft token pruning. In: Proc. European Conference on Computer Vision (ECCV) (2022)

\bibitem{kuznetsova2020openimages}
Kuznetsova, A., Rom, H., Alldrin, N., Uijlings, J., Krasin, I., Pont-Tuset, J., Kamali, S., Popov, S., Malloci, M., Kolesnikov, A., Duerig, T., Ferrari, V.: The open images dataset v4: Unified image classification, object detection, and visual relationship detection at scale. In: International Journal of Computer Vision (IJCV) (2020)

\bibitem{vanlandeghem2023dude}
Landeghem, J.V., Tito, R., Łukasz Borchmann, Pietruszka, M., Józiak, P., Powalski, R., Jurkiewicz, D., Coustaty, M., Ackaert, B., Valveny, E., Blaschko, M., Moens, S., Stanisławek, T.: Document understanding dataset and evaluation (dude). In: Proc. IEEE International Conference on Computer Vision (ICCV) (2023)

\bibitem{lee2026ergo}
Lee, J., Shin, W., Yang, S., Song, K.U., Lim, D., Kim, J., Kim, T.H., Kim, B.K.: Ergo: Efficient high-resolution visual understanding for vision-language models. In: International Conference on Learning Representations (ICLR) (2026)

\bibitem{li2023blip2}
Li, J., Li, D., Savarese, S., Hoi, S.: {BLIP-2:} bootstrapping language-image pre-training with frozen image encoders and large language models. In: Proc. International Conference on Machine Learning (ICML) (2023)

\bibitem{liang2022evit}
Liang, Y., Ge, C., Tong, Z., Song, Y., Wang, J., Xie, P.: Not all patches are what you need: Expediting vision transformers via token reorganizations. In: International Conference on Learning Representations (ICLR) (2022)

\bibitem{Lin_2025}
Lin, W., Feng, Y., Zhu, Y.: <scp>metasapiens:</scp> real-time neural rendering with efficiency-aware pruning and accelerated foveated rendering. In: Proceedings of the 30th ACM International Conference on Architectural Support for Programming Languages and Operating Systems, Volume 1. p. 669–682. ASPLOS ’25, ACM (Mar 2025). \doi{10.1145/3669940.3707227}, \url{http://dx.doi.org/10.1145/3669940.3707227}

\bibitem{lin2023sphinx}
Lin, Z., Liu, C., Zhang, R., Gao, P., Qiu, L., Xiao, H., Qiu, H., Lin, C., Shao, W., Chen, K., Han, J., Huang, S., Zhang, Y., He, X., Li, H., Qiao, Y.: Sphinx: The joint mixing of weights, tasks, and visual embeddings for multi-modal large language models. arXiv preprint arXiv:2311.07575  (2023)

\bibitem{liu2023vsr}
Liu, F., Emerson, G., Collier, N.: Visual spatial reasoning. In: Proc. Annual Meeting of the Association for Computational Linguistics (ACL) (2023)

\bibitem{liu2024llava15}
Liu, H., Li, C., Li, Y., Lee, Y.J.: Improved baselines with visual instruction tuning. In: Proc. IEEE Conference on Computer Vision and Pattern Recognition (CVPR) (2024)

\bibitem{liu2023llava}
Liu, H., Li, C., Wu, Q., Lee, Y.J.: Visual instruction tuning. In: Advances in Neural Information Processing Systems (NeurIPS) (2023)

\bibitem{liu2023revisitingtoken}
Liu, Y., Gehrig, M., Messikommer, N., Cannici, M., Scaramuzza, D.: Revisiting token pruning for object detection and instance segmentation. In: Proc. Winter Conference on Applications of Computer Vision (WACV) (2024)

\bibitem{lu2022scienceqa}
Lu, P., Mishra, S., Xia, T., Qiu, L., Chang, K.W., Zhu, S.C., Tafjord, O., Clark, P., Kalyan, A.: Learn to explain: Multimodal reasoning via thought chains for science question answering. In: Advances in Neural Information Processing Systems (NeurIPS) (2022)

\bibitem{lukanov2021biologically}
Lukanov, H., König, P., Pipa, G.: Biologically inspired deep learning model for efficient foveal-peripheral vision. Frontiers in Computational Neuroscience  \textbf{Volume 15 - 2021} (2021). \doi{10.3389/fncom.2021.746204}, \url{https://www.frontiersin.org/journals/computational-neuroscience/articles/10.3389/fncom.2021.746204}

\bibitem{mathew2021infographicvqa}
Mathew, M., Bagal, V., Tito, R.P., Karatzas, D., Valveny, E., Jawahar, C.V.: Infographicvqa. In: Proc. Winter Conference on Applications of Computer Vision (WACV) (2022)

\bibitem{mathew2021docvqa}
Mathew, M., Karatzas, D., Jawahar, C.V.: Docvqa: A dataset for vqa on document images. In: Proc. Winter Conference on Applications of Computer Vision (WACV) (2021)

\bibitem{min2022pervit}
Min, J., Zhao, Y., Luo, C., Cho, M.: Peripheral vision transformer. In: Advances in Neural Information Processing Systems (NeurIPS) (2022)

\bibitem{mnih2014ram}
Mnih, V., Heess, N., Graves, A., Kavukcuoglu, K.: Recurrent models of visual attention. In: Advances in Neural Information Processing Systems (NeurIPS) (2014)

\bibitem{openai}
{OpenAI}: {Chatgpt} (2025), accessed: 2025-04-05

\bibitem{2024omnilmm}
{OpenBMB}: {MiniCPM-o}. \url{https://github.com/OpenBMB/MiniCPM-o} (2024), accessed: 2024-03-05

\bibitem{plummer2016flickr30k}
Plummer, B.A., Wang, L., Cervantes, C.M., Caicedo, J.C., Hockenmaier, J., Lazebnik, S.: Flickr30k entities: Collecting region-to-phrase correspondences for richer image-to-sentence models. In: Proc. IEEE International Conference on Computer Vision (ICCV) (2015)

\bibitem{qi2024cogcom}
Qi, J., Ding, M., Wang, W., Bai, Y., Lv, Q., Hong, W., Xu, B., Hou, L., Li, J., Dong, Y., Tang, J.: Cogcom: Train large vision-language models diving into details through chain of manipulations. arXiv preprint arXiv:2402.04236  (2024)

\bibitem{qin2025covt}
Qin, Y., Wei, B., Ge, J., Kallidromitis, K., Fu, S., Darrell, T., Wang, X.: Chain-of-visual-thought: Teaching vlms to see and think better with continuous visual tokens. arXiv preprint arXiv:2511.19418  (2025)

\bibitem{rao2021dynamicvit}
Rao, Y., Zhao, W., Liu, B., Lu, J., Zhou, J., Hsieh, C.J.: Dynamicvit: Efficient vision transformers with dynamic token sparsification. In: Advances in Neural Information Processing Systems (NeurIPS) (2021)

\bibitem{rosenholtz2012summary}
Rosenholtz, R., Huang, J., Raj, A., Balas, B.J., Ilie, L.: A summary statistic representation in peripheral vision explains visual search. Journal of Vision  \textbf{12}(4),  14--14 (04 2012). \doi{10.1167/12.4.14}, \url{https://doi.org/10.1167/12.4.14}

\bibitem{ryoo2022tokenlearner}
Ryoo, M.S., Piergiovanni, A., Arnab, A., Dehghani, M., Angelova, A.: Tokenlearner: What can 8 learned tokens do for images and videos? In: Advances in Neural Information Processing Systems (NeurIPS) (2021)

\bibitem{sarch2025vigorl}
Sarch, G., Saha, S., Khandelwal, N., Jain, A., Tarr, M.J., Kumar, A., Fragkiadaki, K.: Grounded reinforcement learning for visual reasoning. In: Advances in Neural Information Processing Systems (NeurIPS) (2025)

\bibitem{shao2024visualcot}
Shao, H., Qian, S., Xiao, H., Song, G., Zong, Z., Wang, L., Liu, Y., Li, H.: Visual cot: Unleashing chain-of-thought reasoning in multi-modal language models. In: Advances in Neural Information Processing Systems (NeurIPS) (2024)

\bibitem{shao2024deepseekmath}
Shao, Z., Wang, P., Zhu, Q., Xu, R., Song, J., Bi, X., Zhang, H., Zhang, M., Li, Y.K., Wu, Y., Guo, D.: Deepseekmath: Pushing the limits of mathematical reasoning in open language models. arXiv preprint arXiv:2402.03300  (2024)

\bibitem{shen2025zoomeye}
Shen, H., Zhao, K., Zhao, T., Xu, R., Zhang, Z., Zhu, M., Yin, J.: Zoomeye: Enhancing multimodal llms with human-like zooming capabilities through tree-based image exploration. In: Conference on Empirical Methods in Natural Language Processing (EMNLP) (2025)

\bibitem{sidorov2020textcaps}
Sidorov, O., Hu, R., Rohrbach, M., Singh, A.: Textcaps: a dataset for image captioning with reading comprehension. In: Proc. European Conference on Computer Vision (ECCV) (2020)

\bibitem{singh2019textvqa}
Singh, A., Natarajan, V., Shah, M., Jiang, Y., Chen, X., Batra, D., Parikh, D., Rohrbach, M.: Towards vqa models that can read. In: Proc. IEEE Conference on Computer Vision and Pattern Recognition (CVPR) (2019)

\bibitem{su2025pixelreasoner}
Su, A., Wang, H., Ren, W., Lin, F., Chen, W.: Pixel reasoner: Incentivizing pixel-space reasoning with curiosity-driven reinforcement learning. In: Advances in Neural Information Processing Systems (NeurIPS) (2025)

\bibitem{su2021roformer}
Su, J., Lu, Y., Pan, S., Wen, B., Liu, Y.: Roformer: Enhanced transformer with rotary position embedding. arXiv preprint arXiv:2104.09864  (2021)

\bibitem{wah2011cub}
Wah, C., Branson, S., Welinder, P., Perona, P., Belongie, S.: Caltech-ucsd birds 200. Tech. Rep. CNS-TR-2011-001, California Institute of Technology (2011)

\bibitem{wang2023cogvlm}
Wang, W., Lv, Q., Yu, W., Hong, W., Qi, J., Wang, Y., Ji, J., Yang, Z., Zhao, L., Song, X., Xu, J., Xu, B., Li, J., Dong, Y., Ding, M., Tang, J.: Cogvlm: Visual expert for pretrained language models. In: Advances in Neural Information Processing Systems (NeurIPS) (2024)

\bibitem{wu2024vstar}
Wu, P., Xie, S.: V*: Guided visual search as a core mechanism in multimodal llms. In: Proc. IEEE Conference on Computer Vision and Pattern Recognition (CVPR) (2024)

\bibitem{xu2025llavacot}
Xu, G., Jin, P., Wu, Z., Li, H., Song, Y., Sun, L., Yuan, L.: Llava-cot: Let vision language models reason step-by-step. In: Proc. IEEE International Conference on Computer Vision (ICCV) (2025)

\bibitem{xu2021evovit}
Xu, Y., Zhang, Z., Zhang, M., Sheng, K., Li, K., Dong, W., Zhang, L., Xu, C., Sun, X.: Evo-vit: Slow-fast token evolution for dynamic vision transformer. In: Proc. AAAI Conference on Artificial Intelligence (AAAI) (2022)

\bibitem{yang2025qwen3}
Yang, A., Li, A., Yang, B., Zhang, B., Hui, B., Zheng, B., Yu, B., Gao, C., Huang, C., Lv, C., Zheng, C., Liu, D., Zhou, F., Huang, F., Hu, F., Ge, H., Wei, H., Lin, H., Tang, J., Yang, J., Tu, J., Zhang, J., Yang, J., Yang, J., Zhou, J., Zhou, J., Lin, J., Dang, K., Bao, K., Yang, K., Yu, L., Deng, L., Li, M., Xue, M., Li, M., Zhang, P., Wang, P., Zhu, Q., Men, R., Gao, R., Liu, S., Luo, S., Li, T., Tang, T., Yin, W., Ren, X., Wang, X., Zhang, X., Ren, X., Fan, Y., Su, Y., Zhang, Y., Zhang, Y., Wan, Y., Liu, Y., Wang, Z., Cui, Z., Zhang, Z., Zhou, Z., Qiu, Z.: Qwen3 technical report. arXiv preprint arXiv:2505.09388  (2025)

\bibitem{yang2025visionthink}
Yang, S., Li, J., Lai, X., Yu, B., Zhao, H., Jia, J.: Visionthink: Smart and efficient vision language model via reinforcement learning. In: Advances in Neural Information Processing Systems (NeurIPS) (2025)

\bibitem{yao2024minicpm}
Yao, Y., Yu, T., Zhang, A., Wang, C., Cui, J., Zhu, H., Cai, T., Li, H., Zhao, W., He, Z., Chen, Q., Zhou, H., Zou, Z., Zhang, H., Hu, S., Zheng, Z., Zhou, J., Cai, J., Han, X., Zeng, G., Li, D., Liu, Z., Sun, M.: Minicpm-v: A gpt-4v level mllm on your phone. arXiv preprint arXiv:2408.01800  (2024)

\bibitem{yarbus2013eye}
Yarbus, A.L.: Eye movements and vision. Springer (2013)

\bibitem{ye2024neural}
Ye, J., Meng, X., Guo, D., Shang, C., Mao, H., Yang, X.: Neural foveated super-resolution for real-time vr rendering. Computer Animation and Virtual Worlds  \textbf{35}(4),  e2287 (2024). \doi{https://doi.org/10.1002/cav.2287}, \url{https://onlinelibrary.wiley.com/doi/abs/10.1002/cav.2287}

\bibitem{yin2022adavit}
Yin, H., Vahdat, A., Alvarez, J., Mallya, A., Kautz, J., Molchanov, P.: Adavit: Adaptive tokens for efficient vision transformer. In: Proc. IEEE Conference on Computer Vision and Pattern Recognition (CVPR) (2022)

\bibitem{yu2025vpt}
Yu, R., Ma, X., Wang, X.: Auto-controlled image perception in mllms via visual perception tokens. In: Proc. IEEE International Conference on Computer Vision (ICCV) (2025)

\bibitem{yu2025docthinker}
Yu, W., Yang, Z., Liu, Y., Bai, X.: Docthinker: Explainable multimodal large language models with rule-based reinforcement learning for document understanding. In: Proc. IEEE International Conference on Computer Vision (ICCV) (2025)

\bibitem{zhang2024llava-reasoner-dpo}
Zhang, R., Zhang, B., Li, Y., Zhang, H., Sun, Z., Gan, Z., Yang, Y., Pang, R., Yang, Y.: Improve vision language model chain-of-thought reasoning. In: Proc. Annual Meeting of the Association for Computational Linguistics (ACL) (2024)

\bibitem{zhang2025adaptivecof}
Zhang, X., Gao, Z., Zhang, B., Li, P., Zhang, X., Liu, Y., Yuan, T., Wu, Y., Jia, Y., Zhu, S.C., Li, Q.: Adaptive chain-of-focus reasoning via dynamic visual search and zooming for efficient vlms. arXiv preprint arXiv:2505.15436  (2025)

\bibitem{zhang2024multimodalchainofthought}
Zhang, Z., Zhang, A., Li, M., Zhao, H., Karypis, G., Smola, A.: Multimodal chain-of-thought reasoning in language models. arXiv preprint arXiv:2302.00923  (2024)

\bibitem{zhao2025uvcot}
Zhao, K., Zhu, B., Sun, Q., Zhang, H.: Unsupervised visual chain-of-thought reasoning via preference optimization. In: Proc. IEEE International Conference on Computer Vision (ICCV) (2025)

\bibitem{zheng2025deepeyes}
Zheng, Z., Yang, M., Hong, J., Zhao, C., Xu, G., Yang, L., Shen, C., Yu, X.: Deepeyes: Incentivizing "thinking with images" via reinforcement learning. In: International Conference on Learning Representations (ICLR) (2025)

\bibitem{zhu2023minigpt4}
Zhu, D., Chen, J., Shen, X., Li, X., Elhoseiny, M.: Minigpt-4: Enhancing vision-language understanding with advanced large language models. In: International Conference on Learning Representations (ICLR) (2024)

\bibitem{zhu2016visual7w}
Zhu, Y., Groth, O., Bernstein, M., Fei-Fei, L.: Visual7w: Grounded question answering in images. In: Proc. IEEE Conference on Computer Vision and Pattern Recognition (CVPR) (2016)

\end{thebibliography}

\end{document}